\documentclass[10pt,journal,compsoc]{IEEEtran}
\usepackage[nocompress]{cite}

\usepackage{graphicx}

\usepackage{subfigure}
\usepackage{caption}

\usepackage{multirow}
\usepackage{multicol}
\usepackage{array}
\usepackage{enumitem}
\usepackage{multirow}
\usepackage{diagbox}

\usepackage{dsfont}
\usepackage[pagebackref=true,breaklinks=true,letterpaper=true,colorlinks,bookmarks=false,urlcolor=magenta]{hyperref}
\usepackage{amsmath}
\usepackage{amsfonts}
\ifCLASSINFOpdf
\else
\fi

\begin{document}
%
\title{HUMBI: A Large Multiview Dataset of Human Body Expressions and Benchmark Challenge}
%
%
%
%

\author{Jae Shin Yoon, Zhixuan Yu, Jaesik Park, Hyun Soo Park
\IEEEcompsocitemizethanks{\IEEEcompsocthanksitem J. S. Yoon, Z. Yu, and H. S. Park are with University of Minnesota.\protect\\
E-mail: \{jsyoon, yu000064, hspark\}@umn.edu
\IEEEcompsocthanksitem J. Park is with POSTECH. E-mail: jaesik.park@postech.ac.kr
}
}

\definecolor{RedColor}{rgb}{0.7,0,0} 
\definecolor{GreenColor}{rgb}{0,0.4,0} 
\definecolor{BlueColor}{rgb}{0,0,0.8} 
\definecolor{CyanColor}{rgb}{0,1,1} 
\newcommand{\new}[1]{{\color{RedColor} #1 $\qed$}}

\IEEEtitleabstractindextext{%
\begin{abstract}
This paper presents a new large multiview dataset called HUMBI for human body expressions with natural clothing. The goal of HUMBI is to facilitate modeling view-specific appearance and geometry of five primary body signals including gaze, face, hand, body, and garment from assorted people. 107 synchronized HD cameras are used to capture 772 distinctive subjects across gender, ethnicity, age, and style. With the multiview image streams, we reconstruct the geometry of body expressions using 3D mesh models, which allows representing view-specific appearance. We demonstrate that HUMBI is highly effective in learning and reconstructing a complete human model and is complementary to the existing datasets of human body expressions with limited views and subjects such as MPII-Gaze, Multi-PIE, Human3.6M, and Panoptic Studio datasets. Based on HUMBI, we formulate a new benchmark challenge of a pose-guided appearance rendering task that aims to substantially extend photorealism in modeling diverse human expressions in 3D, which is the key enabling factor of authentic social tele-presence. HUMBI is publicly available at \url{http://humbi-data.net}.


\end{abstract}



%

\begin{IEEEkeywords}
Human behavioral imaging, multiview dataset, 3D geometry and appearance
\end{IEEEkeywords}

}

\maketitle
\section{Introduction}\label{sec:intro}
Humans possess a quintessential sensitivity to effortlessly read invisible internal states of others, e.g., intent, emotion, and attention, through every nuance of their body expressions, including gaze, face, and gestures. It is impossible, therefore, to enable authentic social presence in a virtual space without conveying photorealistic models of such body expressions. This is, however, extremely challenging because it requires decoding complex physical interactions between texture, geometry, illumination, and viewpoint (e.g., translucent skins, tiny wrinkles, and reflective fabric) from an image of a subject.



Recently, pose- and view-specific models by making use of a copious capacity of neural encoding~\cite{lombardi2019neural, meka2020deep} substantially extend the expressibility of existing linear models~\cite{cootes:2001}. So far, these models have been constructed by a sequence of the detailed scans of a target subject using dedicated camera infrastructure (e.g., multi-camera systems~\cite{joo_cvpr_2014,Bee10,Wenger:2005}), i.e., they are subject-specific which is not generalizable to other subjects. Looking ahead, we would expect a new versatile model that is applicable to the general appearance of assorted people by eliminating the requirement of the massive scans for every target subject.

\begin{figure}[t]
    \begin{center}
    
        \includegraphics[width=0.26\textwidth,trim={0 0 13.5cm 0},clip]{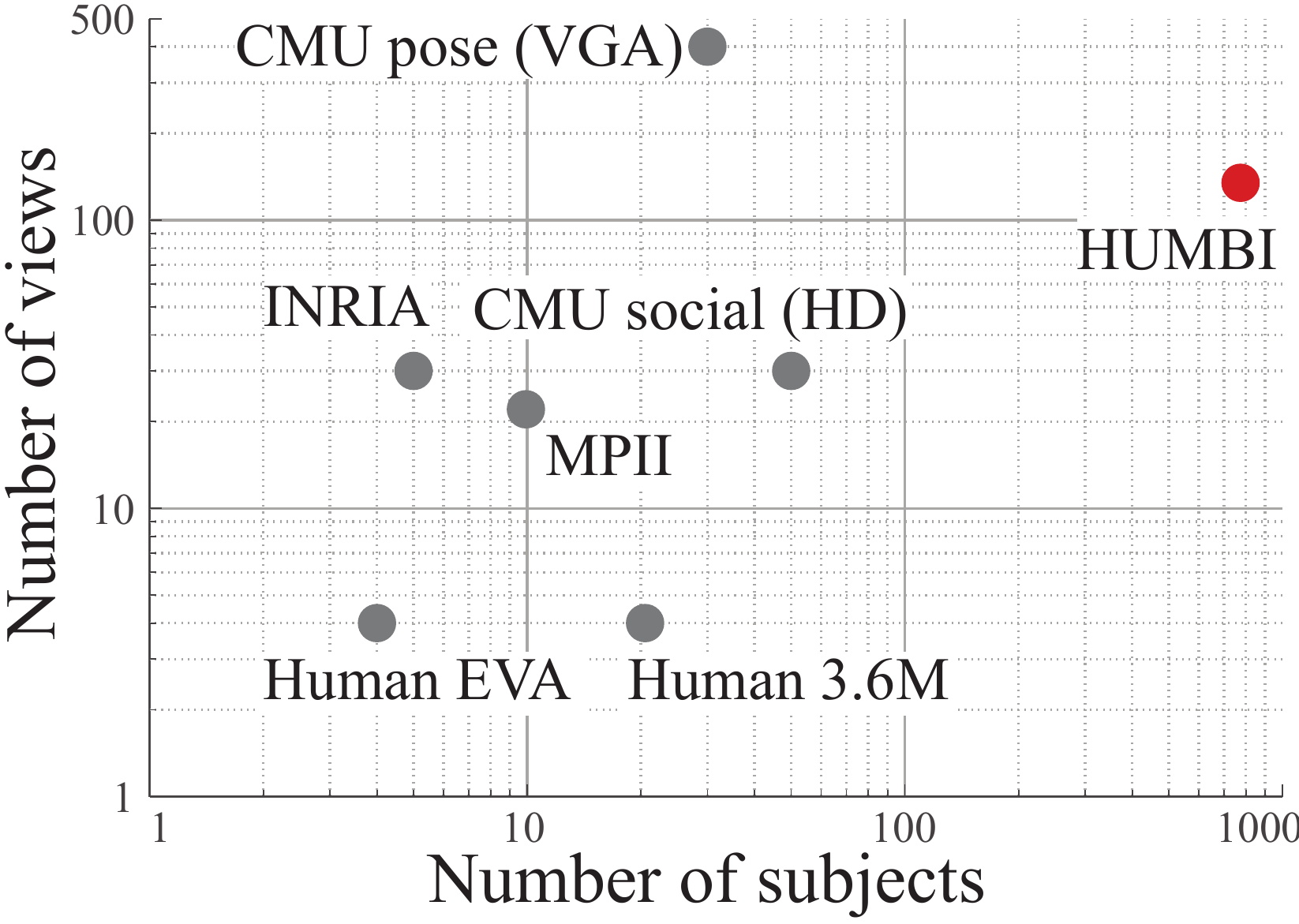}\hspace{3mm}
        \includegraphics[width=0.19\textwidth,trim={18cm 0 0  0},clip]{position.pdf}
    \end{center}
    \vspace{-2mm}
    \caption{We present HUMBI that pushes towards two extremes: views and subjects. Comparing to existing datasets such as CMU Panoptic Studio~\cite{joo:2015,joo:2019}, MPII~\cite{Dyna:SIGGRAPH:2015,pons2017clothcap}, and INRIA~\cite{Knossow:2008}, HUMBI presents the unprecedented scale visual data measured by 107 HD cameras that can be used to learn the detailed appearance and geometry of five elementary human body expressions for 772 distinctive subjects.}
    \label{fig:position}
    \vspace{-4mm}
\end{figure}

\begin{figure*}[t]
    \begin{center}
        	\includegraphics[trim=0 90mm 0 48mm, clip, width=\textwidth]{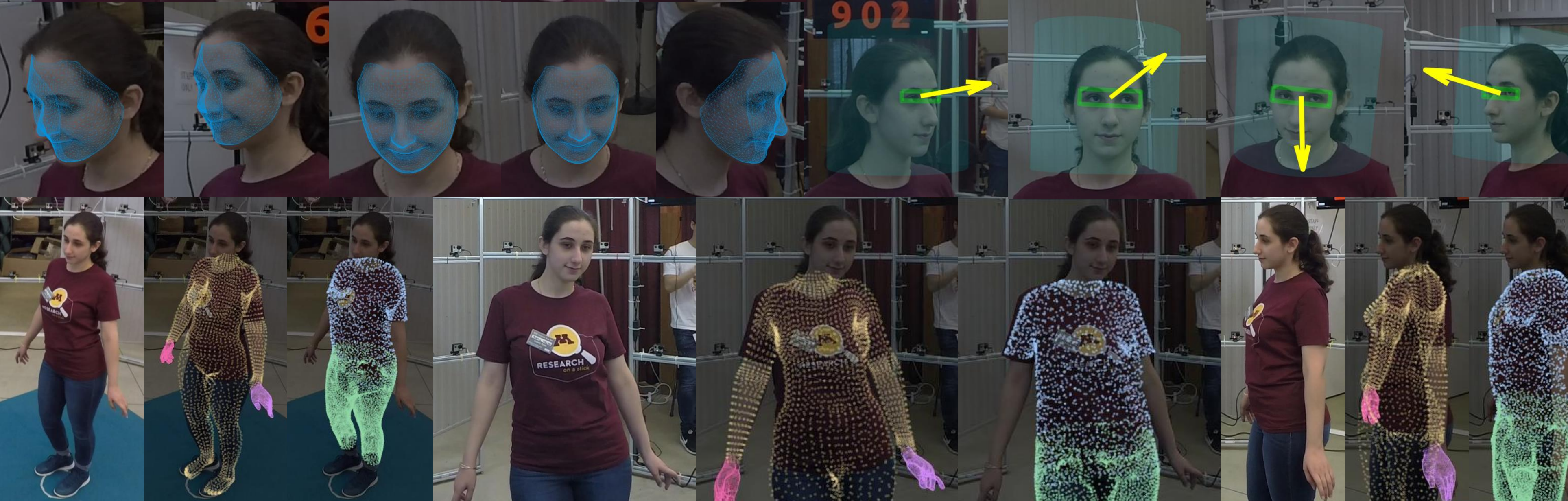}
    \end{center}
    \vspace{-4mm}
\captionof{figure}{We present a new large multiview dataset dataset of human body expressions called HUMBI. 
We focus on five elementary expressions: face (blue), gaze (yellow), hand (pink and purple), body (light orange), and garment including top (light blue) and bottom (light green).}
	\label{fig:teaser_big}
\end{figure*}

    
Among many factors, what are the core resources to build such a generalizable model? We argue that the data that can span an extensive range of appearances from numerous shapes and identities are prerequisites. To validate our conjecture, we present a new dataset of human body expressions called~\textit{HUMBI} (HUman Multiview Behavioral Imaging) that pushes to two extremes: views and subjects. As shown in Fig.~\ref{fig:position}, HUMBI is composed of 772 distinctive subjects with natural clothing across diverse age, gender, ethnicity, and style captured by 107 HD synchronized cameras (68 cameras facing at frontal body). This poses unprecedented diversity of visual data that are ideal for modeling generalizable geometry and appearance, which is not presented in existing datasets including CMU~\cite{joo:2015,joo:2019} and MPII~\cite{Dyna:SIGGRAPH:2015,pons2017clothcap} as shown in Fig.~\ref{fig:position}. We focus on five elementary human body expressions as shown in Fig~\ref{fig:teaser_big}.: gaze, face, hand, body, and garment.

The main properties of HUMBI are summarized below. 
(1) Complete: it captures the total human appearance, including gaze, face, hand, foot, body, and garment to represent holistic body signals~\cite{Joo:2018}, e.g., perceptual asynchrony between the face and hand movements. 
(2) Dense: 107 HD cameras create a dense light field that observes the minute body expressions with minimal self-occlusion. This dense light field allows us to model precise appearance as a function of view~\cite{LOMBARDI:2018}. 
(3) Natural: the subjects are all voluntary participants (no actor/actress/student/researcher). Their activities are loosely guided by performance instructions, which generates natural body expressions. (4) Diverse: it captures 772 distinctive subjects with diverse clothing styles, skin colors, time-varying geometry of gaze/face/body/hand, and range of motion. (5) Fine: with multiview HD cameras, we reconstruct the high fidelity 3D model using 3D meshes, which allows representing view-specific appearance in its canonical atlas.  (6) Effective: we show that vanilla convolutional neural networks (CNN) designed to learn view-invariant 3D pose geometry from HUMBI quantitatively outperform the counterpart models trained by existing datasets. More importantly, we show that it is \textit{complementary} to such datasets, i.e., the trained models can be substantially improved by combining with these datasets.

Owing to these properties, HUMBI is an ideal dataset to evaluate the ability of modeling human appearance and geometry as shown in Fig.~\ref{fig:motivation}. To measure such ability, we formulate a novel benchmark challenge on a pose-guided appearance rendering task: given a single view image of a person, render the person appearance from other views and poses. HUMBI offers the ground truth of this challenging task where the performance of the approaches can be precisely characterized. We validate the feasibility of the benchmark challenge using the state-of-the-art rendering methods~\cite{ma2017pose,zhu2019progressive,tang2019cycle,ren2020deep}. 

\textbf{Contribution} This paper is built upon an earlier version~\cite{yu2020humbi} which includes (1) a new large multiview dataset of diverse natural expressions of assorted people for view-specific appearance; (2) a re-configurable camera system design that allows precise capture in public spaces; (3) cross-dataset evaluation that characterizes strength and complementary property of HUMBI. The core contributions newly made from this paper are as follows: (4) We present a new benchmark challenge for the task of pose-guided human rendering. We evaluate and analyze recent methods on this new benchmark. The dataset, pre-computed results, and leaderboard are available at our project page; (5) We provide more detailed description for our performance capture system including its new illustration, camera calibration, and procedure; (6) We introduce 3D semantic point clouds dataset, which allows modeling category-specific geometry.

\begin{table*}[h]
\centering
\scriptsize
\resizebox{0.995\linewidth}{!}{
\begin{tabular}{l|l|l|l|l|l|l|l}
\hline
Dataset & \# of subjects & Measurement method & Gaze & Face & Hand & Body & Cloth\\
\hline
\hline
Columbia Gaze~\cite{smith:2013} & 56 & 5 cameras & \checkmark (fixed) &  & &  & \\
UT-Multiview~\cite{sugano:2014} & 50 & 8 cameras & \checkmark (fixed) &  & &  & \\
Eyediap~\cite{Mora:2014} & 16 & 1 depth camera and 1 HD camera & \checkmark (free) &  & &  & \\
MPII-Gaze~\cite{zhang15_cvpr} & 15 & 1 camera & \checkmark (free)  &  & &  & \\
RT-GENE~\cite{fischer2018rt} & 15 & eyetracking device & \checkmark (free)  &  & &  & \\
ETH-XGaze~\cite{zhang2020eth} & 110 & 18 SLR cameras & \checkmark (free)  &  & &  & \\
\hline
CMU Multi-PIE~\cite{Gross2009} & 337 & 15 cameras &  & \checkmark & &  & \\
3DMM~\cite{blanz:2003} & 200 & 3D scanner &  & \checkmark & &  & \\
BFM~\cite{bfm09} & 200 & 3D scanner &  & \checkmark & &  & \\
ICL~\cite{booth2018large} & 10,000 & 3D scanner &   & \checkmark & &  & \\
FaceScape~\cite{yang2020facescape}&938 & 68 DSLR cameras & & \checkmark & & \\

\hline
NYU Hand~\cite{tompson:2014}& 2 (81K samples) & Depth camera &   &  & \checkmark &  & \\
HandNet~\cite{Wetzler:2016}& 10 (213K samples) & Depth camera and magnetic sensor &   &  & \checkmark &  & \\
BigHand 2.2M~\cite{yuan:2017}& 10 (2.2M samples) & Depth camera and magnetic sensor &   &  & \checkmark &  & \\
RHD~\cite{zimmermann2017learning}& 20 (44K samples) &  N/A (synthesized) &   &  & \checkmark &  & \\
STB~\cite{zhang2017hand}& 1 (18K samples) &  1 pair of stereo cameras &   &  & \checkmark &  & \\
FreiHand~\cite{Freihand2019}& N/A (33K samples) & 8 cameras &   &  & \checkmark &  & \\
\hline
CMU Mocap & $\sim$100& Marker-based & &  & & \checkmark & \\ 
CMU Skin Mocap~\cite{park:2006} & $<$10 & Marker-based & & \checkmark & & \checkmark & \\
INRIA~\cite{Knossow:2008} & N/A & Markerless (34 cameras) & & & & \checkmark & \checkmark (natural)\\
Human EVA~\cite{sigal2010humaneva} & 4 & Marker-based and Markerless (4-7 cameras) & & & & \checkmark & \\
Human 3.6M~\cite{h36m_pami} & 11 & Markerless (depth camera and 4 HD cameras) & & & & \checkmark & \\
Panoptic Studio~\cite{Joo_2017_TPAMI,simon2017hand} & $\sim$100 & Markerless (31 HD and 480 VGA cameras) &  &  & \checkmark & \checkmark  & \\
Dyna~\cite{Dyna:SIGGRAPH:2015} & 10 & Markerless (22 pairs of stereo cameras) &  &  & &  \checkmark & \\
ClothCap~\cite{pons2017clothcap} & 10 & Markerless (22 pairs of stereo cameras) &  &  &  & & \checkmark (natural)\\
BUFF~\cite{Zhang_2017_CVPR} & 5 & Markerless (22 pairs of stereo cameras) &  &  &  & \checkmark & \checkmark (natural) \\
3DPW~\cite{vonMarcard2018} & 7 & Marker-based (17 IMUs) and Markerless (1 camera + 3D scanner) &  &  &  & \checkmark & \checkmark (natural)  \\
TNT15~\cite{vonPon2016a} & 4 & Marker-based (10 IMUs) and Markerless (8 cameras + 3D scanner)  &  &  &  & \checkmark & \\
D-FAUST\cite{dfaust:CVPR:2017} & 10 & Markerless (22 pairs of stereo cameras) &  &  &  & \checkmark & \\
\hline
\hline
HUMBI & 772  & Markerless (107 HD cameras) & \checkmark (free) & \checkmark & \checkmark & \checkmark  & \checkmark (natural)\\
\hline
\end{tabular}}
\caption{Human body expression datasets. HUMBI is designed to capture whole body expressions of 772 subjects from synchronized 107 multiple cameras, which forms tera-scale multiview visual data. Along with images, HUMBI provides the 3D mesh models for each body expression and 3D point clouds as a reference. Note that, HUMBI is not intended to provide 3D ground truth for fine-grained geometry (as opposed to dedicated multimodal capture, e.g., ClothCap~\cite{pons2017clothcap} and DeepFasion3D~\cite{zhu2020deep}) while some may find useful as a pseudo-ground truth. The images and the reconstruction results will encourage future works to advance the algorithms for generalizable appearance modeling and high-fidelity 3D reconstruction of dynamic humans.}
\label{table:dataset}
\end{table*}

\section{Related Work}\label{sec:related}


We review the existing datasets for modeling human body expressions of gaze, face, hand, body. These datasets are summarized in Table~\ref{table:dataset}. We further discuss key existing approaches and datasets of human appearance rendering.

\noindent\textbf{Gaze} Columbia Gaze dataset~\cite{smith:2013} and UT-Multiview dataset~\cite{sugano:2014} were captured in a controlled environments where the head poses are fixed. Eyediap dataset~\cite{Mora:2014} captured gaze movement while head is in motion, providing natural gaze movements. MPII-Gaze dataset~\cite{zhang15_cvpr} collected in-the-wild gaze data from the cameras in the tablets, constituting 214K images across 15 subjects. This contains a variety of appearance and illumination. RT-GENE dataset~\cite{fischer2018rt} took a step further by measuring free-ranging point of regard where the ground truth was obtained by using motion capture of mobile eye-tracking glasses. Recently, ETH-XGaze~\cite{zhang2020eth} captured the high resolution eye images from 110 participants under various illumination conditions.

\noindent\textbf{Face} 3D Morphable Model (3DMM)~\cite{blanz:2003} was constructed by 3D scans of large population to model the complex geometry and appearance of human faces. For instance, 3D faces were reconstructed by leveraging facial landmarks~\cite{jourabloo2016large, sagonas2016300, le2012interactive, sagonas2013semi,belhumeur2013localizing}, and dense face mesh~\cite{tewari17MoFA, feng2018prn}. Notably, 3DMM is fitted to 60K samples from several face alignment datasets~\cite{messer1999xm2vtsdb,sagonas2013300,zhou2013extensive,belhumeur2013localizing, zhu2012face} to create the 300W-LP dataset~\cite{zhu2016face}. For facial appearance, a deep appearance model~\cite{LOMBARDI:2018} introduces view-dependent appearance using a conditional variational autoencoder, which outperforms linear active appearance model~\cite{cootes:2001}. Recently, a large-scale detailed 3D face model with pore-level facial geometry and texture are captured from 68 DSLR cameras~\cite{yang2020facescape}.

\noindent\textbf{Hand} Dexterous hand manipulation frequently introduces self-occlusion, which makes building a 3D hand pose dataset challenging. A depth image that provides trivial hand segmentation in conjunction with tracking has been used to establish the ground truth hand pose~\cite{tompson:2014,sun:2015,tang:2014,supancic:2015}. However, such approaches still require intense manual adjustments. This challenge was addressed by making use of graphically generated hands~\cite{mueller2018ganerated,zimmermann2017learning,mueller2017real}, which may introduce a domain gap between real and synthetic data. For real data, an auxiliary input such as magnetic sensors was used to precisely measure the joint angle and recover 3D hand pose using forward kinematics~\cite{Wetzler:2016,yuan:2017}. Notably, a multi-camera system has been used to annotate hands using 3D bootstrapping~\cite{simon2017hand}, which provided the hand annotations for RGB data. FreiHAND\cite{Freihand2019} leveraged MANO\cite{MANO:SIGGRAPHASIA:2017} mesh model to represent dense hand pose.
\noindent\textbf{Body} Markerless motion capture is a viable solution to measure dense human body expression at high resolution. For example, multi-camera systems have been used to capture a diverse set of body poses, e.g., actors and actresses perform a few scripted activities such as drinking, answering cellphone, and sitting~\cite{h36m_pami, sigal2010humaneva}. Natural 3D human behaviors were captured in the midst of the role-playing of social events from a multiview system~\cite{Joo_2017_TPAMI} while those events inherently involve with a significant occlusion by people or objects, which prohibits from modeling a complete human body. Further, a 4D scanner~\cite{dfaust:CVPR:2017,Dyna:SIGGRAPH:2015} enabled high resolution body capture to construct a parametric human models, e.g., SMPL~\cite{loper2015smpl}. Notably, image-to-surface correspondences on 50K COCO images~\cite{lin2014microsoft} enabled modeling humans from a single view image~\cite{Lassner:UP:2017}, and optimizing 3D pose estimation with human dynamics priors such as foot contact states and local rigid body transformation improve the physically plausibility in motion capture from a video of a single camera~\cite{shimada2020physcap}. Further, rendering of human model in images could alleviate annotation efforts~\cite{varol17_surreal}.

\begin{figure}[t]
    \begin{center}
    
        \includegraphics[width=0.48\textwidth,clip]{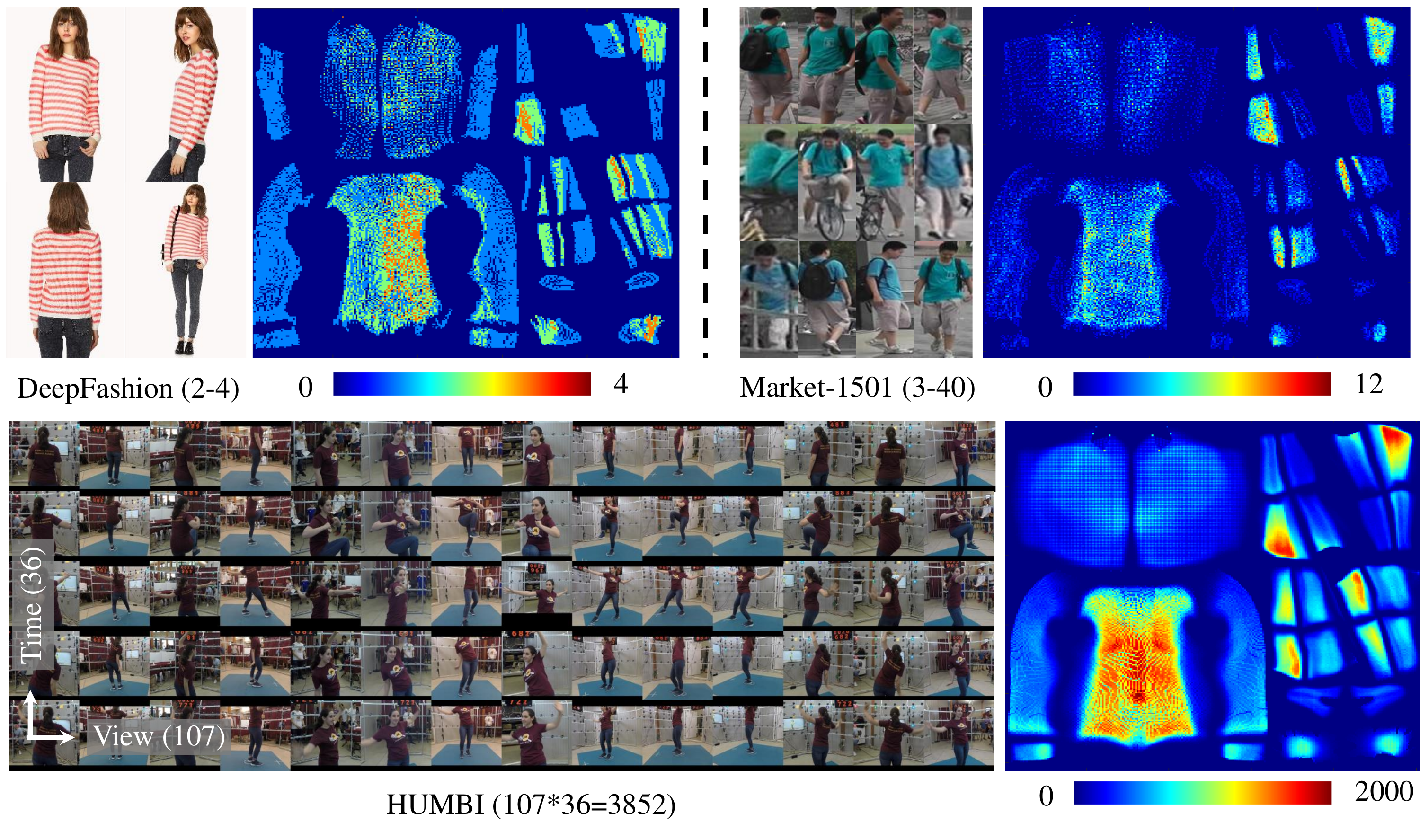}
    \end{center}
    \vspace{-4mm}
    \caption{\small The existing datasets (Deepfashion~\cite{liuLQWTcvpr16DeepFashion} and Market-1501~\cite{zheng2015scalable}) are designed for the task of person re-identification and fashion retrieval, which includes the images captured from limited viewpoints. On the other hand, HUMBI provides images captured from dense camera array, which is ideal to develop and evaluate a human rendering model. The body surface visibility for each dataset is visualized~\cite{guler2018densepose}. The colormap describes the number of cameras visible at each pixel.}
    \label{fig:motivation}
\end{figure}

\begin{figure*}[t]
	\begin{center}
		\includegraphics[width=1\textwidth]{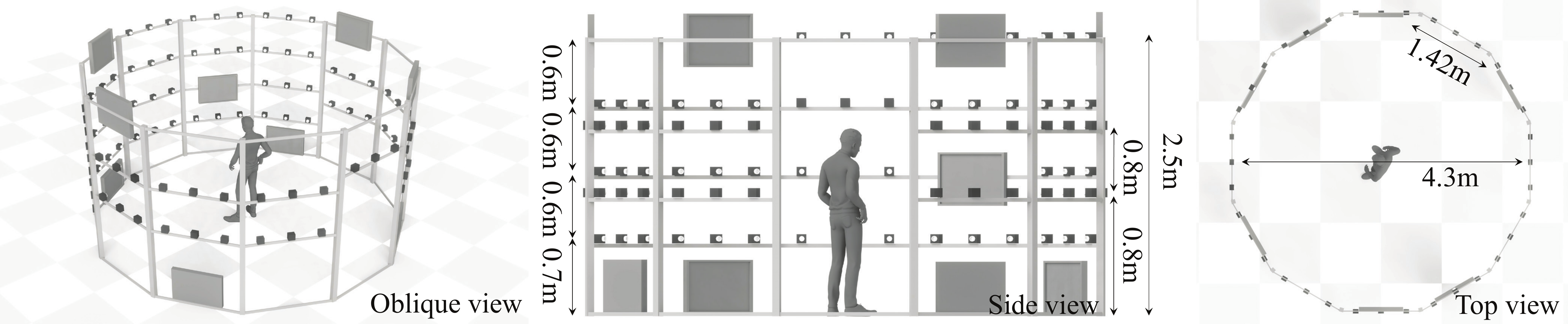}
	\end{center}
	\vspace{-3mm}
    \caption{Re-configurable dodecagon design and its dimension of the multi-camera system.}
    \label{fig:system}
	\vspace{-3mm}
\end{figure*}

\noindent\textbf{Garment} Previous works have proposed to capture the dataset for the natural clothing deformation in response to human body movement. Clothing regions were segmented in 3D using multiview reconstruction~\cite{white2007capturing,bradley2008markerless}. To ensure the same topology when segmenting the clothing out from 3D reconstruction, the SMPL body model can be used to parameterize clothing motion, which produces physically plausible clothing geometry while preserving wrinkle level details~\cite{pons2017clothcap}. Zhu et al.~\cite{zhu2020deep} introduces a large-scale repository for 3D clothing models reconstructed from real garments with limited poses (i.e., a single pose per garment). Skinning is used to deform the clothed model, resulting in diverse posed model~\cite{bertiche2020cloth3d,jiang2020bcnet}. While these 3D modeling makes it possible to render the images of dressed humans from any viewpoints with various synthetic appearance, there still exists a domain gap between the synthesized and real images.

%

\noindent\textbf{Modeling Human Body Expressions} A large body of previous work learned the 3D geometry of human body expressions using multiview image datasets (e.g., a 3D model seen by a pair of views). Some works leveraged a multiview dataset to learn a person-specific 3D human model from a small number of images, which eliminates the requirement of a large multiview camera system ~\cite{habermann2019livecap,habermann2020deepcap,bhatnagar2019mgn,yoon2021neural}. 
The 3D cloth deformation was represented as a function of body shape and pose by learning the nonlinear spatial relationship between the fitted body model and cloth motion~\cite{santesteban2019learning,patel2020tailornet,tiwari2020sizer,gundogdu2020garnet++}.
%
This complex relationship can be encoded in an implicit function~\cite{chibane2020implicit}.  
The 3D point clouds reconstructed from multiview images have been used to infer surface correspondences of the scanned human using a parametric body model~\cite{bhatnagar2020loopreg}, which enables a personalized animatable 3D model~\cite{bhatnagar2020combining,alldieck2019learning,alldieck2018video}.


\noindent\textbf{Pose-guided Human Rendering} Rendering detailed appearance of human body expressions is a challenging task. Modern designs of rendering approaches leverage deep neural networks that offer strong generalizability across diverse identities, shapes, illuminations, and viewpoints. Various generative models have been used to render human images using shape priors, e.g., keypoints~\cite{tang2019cycle,tang2020multi}, heatmap~\cite{zhu2019progressive}, and shape mask~\cite{esser2018variational,ma2017pose}. A person semantic map, i.e., the semantic labels of human body parts, can provide an additional cue to generate photo-realistic images~\cite{song2019unsupervised, han2019clothflow}, and the coarse-to-fine generation can effectively and efficiently combine low and high resolution features to model the details, e.g., complex cloth textures and wrinkles~\cite{yang2020towards}. To provide an explicit guidance to transfer the pixels from a reference view to a target, the dense correspondences are estimated using the priors of flow prediction~\cite{ren2020deep}, person deformation~\cite{siarohin2018deformable} and 3D body model~\cite{li2019dense, Sarkar2020,yoon2021pose}. 
Some existing works explicitly reconstruct animatable 3D human models from a single image for free-viewpoint rendering of a person from different body poses. For example, the geometry and appearance of animatable human models are jointly predicted by regressing a partial texture map in the UV coordinates~\cite{alldieck2019tex2shape,lazova2019360}. Liu et al.~\cite{liu2020neural} enhances the dynamic properties by modeling pose-dependent appearance such as wrinkles and garment deformation from the UV texture map. Graphics knowledge such as rigging and skinning can be combined to animated the 3D human reconstructed from a single image~\cite{weng2019photo}.

\noindent\textbf{Our Approach} Unlike existing datasets focusing on each body expressions, HUMBI is designed to span geometry and appearance of total body expressions from a number of distinctive subjects using a dense camera array. Our tera-scale multiview visual data provide a new opportunity to generalize pose- and view-specific appearance, which can be evaluated by our proposed benchmark.
Existing datasets~\cite{liuLQWTcvpr16DeepFashion,zheng2015scalable} are complementary to HUMBI for the benchmark challenge of a pose-guided human rendering task while their quality assessment can be limited to a few sparse viewpoints as shown in Fig.~\ref{fig:motivation}. HUMBI provides the first multiview benchmark dataset which is composed of the images of 140 people (100 for training, 40 for testing) with 107 views and 36 body poses (3,852 images per person and ~540K in total).

\begin{figure*}[t]
	\begin{center}
		\includegraphics[trim={0 0 0 0},clip,width=\textwidth]{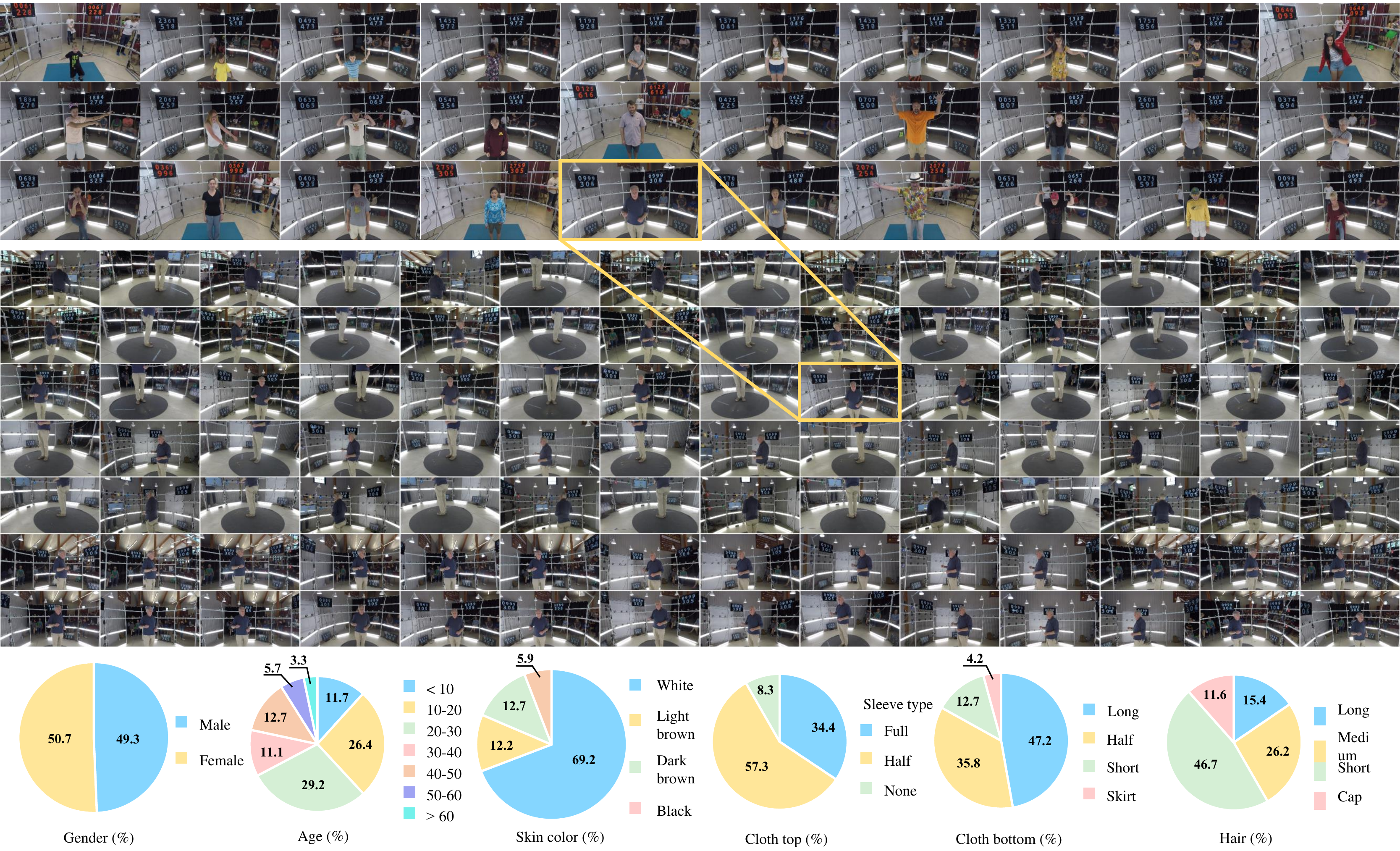}
	\end{center}
    \caption{(Top and bottom) HUMBI includes 772 distinctive subjects across gender, ethnicity, age, clothing style, and physical condition, which generates diverse appearance of human expressions. (Middle) For each subject, 107 HD cameras capture her/his expressions including gaze, face, hand, body, and garment.}
    \label{fig:subject}
\end{figure*}

\section{Multi-camera Imaging System}\label{sec:system}

We design a re-configurable multi-camera system that was deployed in public events including Minnesota State Fair and James Ford Bell Museum of Natural History at the University of Minnesota. 


\noindent\textbf{Hardware} The stage is made of a re-configurable dodecagon frame with 4.2 m diameter and 2.5 m height using T-slot structural framing (80/20 Inc.) where the baseline between adjacent cameras is approximately 10$^\circ$ (22 cm) as shown in Fig.~\ref{fig:system}. The stage is encircled by 107 GoPro HD cameras (38 HERO 5 BLACK Edition and 69 HERO 3+ Silver Edition), one LED display for an instructional video, eight LED displays for video synchronization, and additional lightings. Among 107 cameras, 69 cameras are uniformly placed along the two levels of the dodecagon arc (0.8 m and 1.6 m) for body and clothing, and 38 cameras are place over the frontal hemisphere for face and gaze. 

\noindent\textbf{Instructional Performance Guidance}
To guide the movements of the participants, we create four instructional videos ($\sim$2.5 minutes). Each video is composed of four sessions. (1) Gaze: subjects were asked to find and look at the requested number tag posted on the camera stage; (2) Face: subjects were asked to follow 20 distinctive dynamic facial expressions (e.g., eye rolling, frowning, and jaw opening); (3) Hand: subjects were asked to follow a series of American sign languages (e.g., counting one to ten, greeting, and daily used words);  (4) Body and garment: subjects were asked to follow range of motion, which allows them to move their full body and to follow slow and full speed dance performances curated by a professional choreographer.

\noindent\textbf{Synchronization and Calibration} We manually synchronize 107 cameras using LED displays. The maximum synchronization error is up to 15 ms. We use the COLMAP~\cite{schoenberger2016sfm} software to calibrate camera intrinsics and extrinsics parameters and upgrade the extrinsic parameters to a metric space: the scale is corrected using physical distance between cameras, the origin is translated to the center of the stage, and the orientation is mapped such that its \textit{y}-axis is aligned with the surface normal of the ground plane.

\section{HUMBI}\label{sec:humbi}


HUMBI is composed of 772 distinctive subjects, where each subject includes five elementary body expressions: gaze, face, hand, body, and garment. Notable subject statistics includes: evenly distributed gender (50.7\% female; 49.3\% male); a wide range of age groups (26\% of teenagers, 29\% of 20s, and 11\% of 30s); diverse skin colors (black, dark brown, light brown, and white); various styles of clothing (dress, short-/long-sleeve t-shirt, jacket, hat, and short-/long-pants) as shown in Fig.~\ref{fig:subject}. 



\begin{figure}[t]
	\begin{center}
		\includegraphics[width=3.3in]{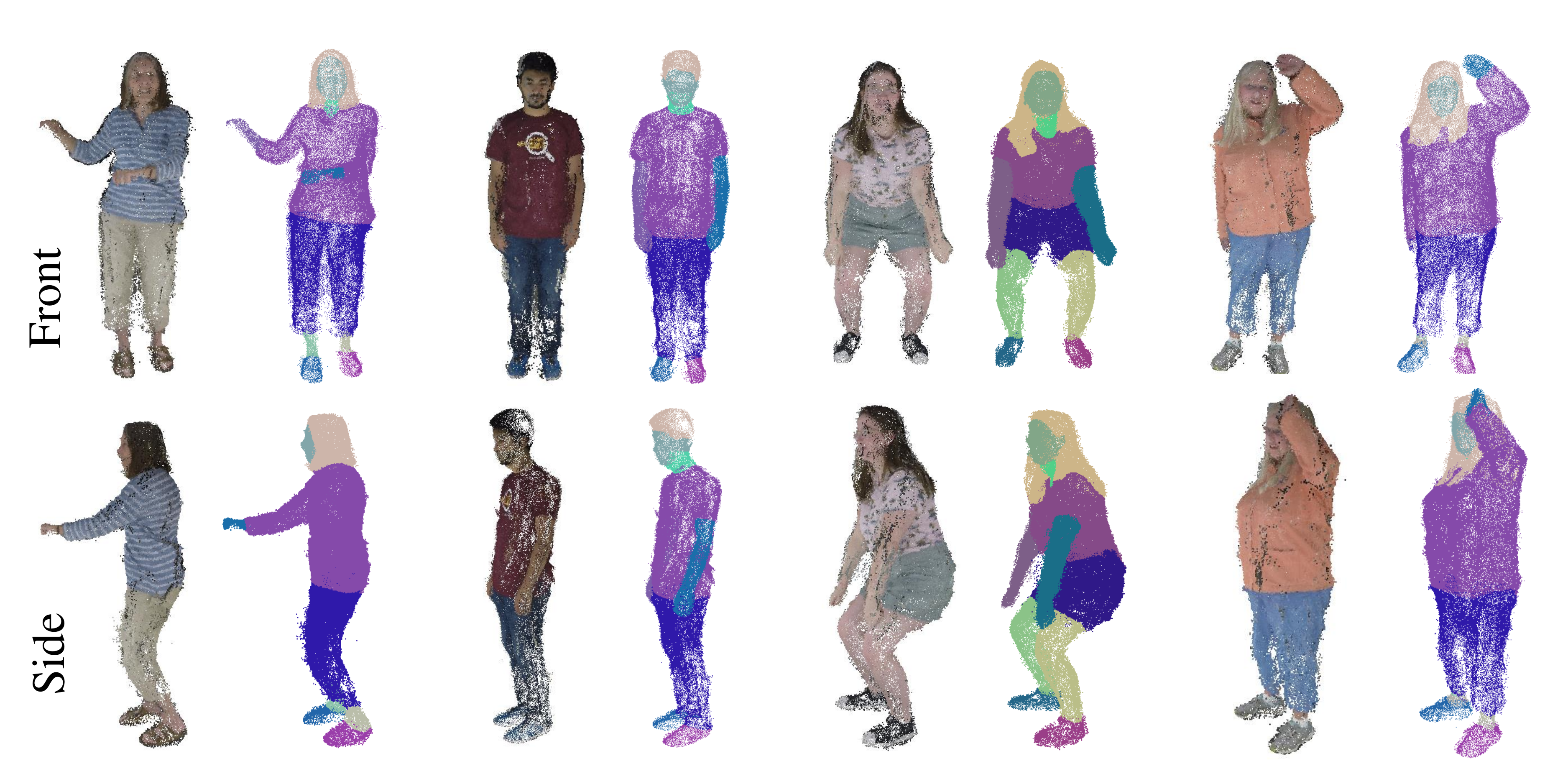}
	\end{center}
    \vspace{-2mm}
    \caption{HUMBI 3D semantic point clouds dataset. We use multiview 3D reconstruction~\cite{schoenberger2016sfm} and segmentation~\cite{yoon20173d} methods to obtain 3D point clouds for each semantic body parts.}
    \label{fig:semantics}
    \vspace{-2mm}
\end{figure}

\noindent\textbf{Notation} We denote our representation of human body expressions as follows:
\begin{itemize}
    \item[\tiny$\bullet$] 3D keypoints: $\mathcal{K}$.
    \item[\tiny$\bullet$] 3D vertices: $\mathbf{V}$.
    \item[\tiny$\bullet$] 3D occupancy map: $\mathcal{O}:\mathds{R}^3\rightarrow \{0,1\}$ that takes as input 3D voxel coordinate and outputs binary occupancy.
    \item[\tiny$\bullet$] Appearance map: $\mathcal{A}:\mathds{R}^2\rightarrow [0,1]^3$ that takes as input atlas coordinate (UV) and outputs normalized RGB values.
\end{itemize}
In the subsequent subsections, we use an abuse of notation for the common variables across body expressions. For example, $\mathcal{K}$ in Sec.~\ref{humbi:face} means $\mathcal{K}_{\rm face}$ while $\mathcal{K}_{\rm hand}$ in Sec.~\ref{humbi:hand}.

\noindent\textbf{3D Body Keypoint Reconstruction} Given a set of synchronized and undistorted multiview images, we detect 2D keypoints of face, hand, body (including feet)~\cite{cao2017realtime}. Using these keypoints, we triangulate 3D keypoints with RANSAC~\cite{Fischler:1981} followed by the non-linear refinement by minimizing reprojection error~\cite{hartley:2004}\footnote{When multiple persons are detected, we use a geometric verification to identify each subject.}. This results in 3D body keypoints on face ($\mathcal{K}_{\rm face}\in\mathds{R}^{3\times 66}$), hands ($\mathcal{K}_{\rm hand}\in\mathds{R}^{3\times 21}$), and body including feet ($\mathcal{K}_{\rm body}\in\mathds{R}^{3\times 25}$).

\noindent\textbf{3D Semantic Point Clouds Reconstruction} We provide dense 3D point cloud reconstructed by a multiview stereo~\cite{schoenberger2016sfm}. Given the 3D reconstructed point clouds, we use a multiview inference~\cite{yoon20173d} to assign a semantic label (20 categories specifying hand, hair, pants, etc). This allows modeling part-specific geometry such as garment and face as shown in Fig.~\ref{fig:semantics}.


\begin{figure}[t]
	\begin{center}
		\includegraphics[width=0.45\textwidth]{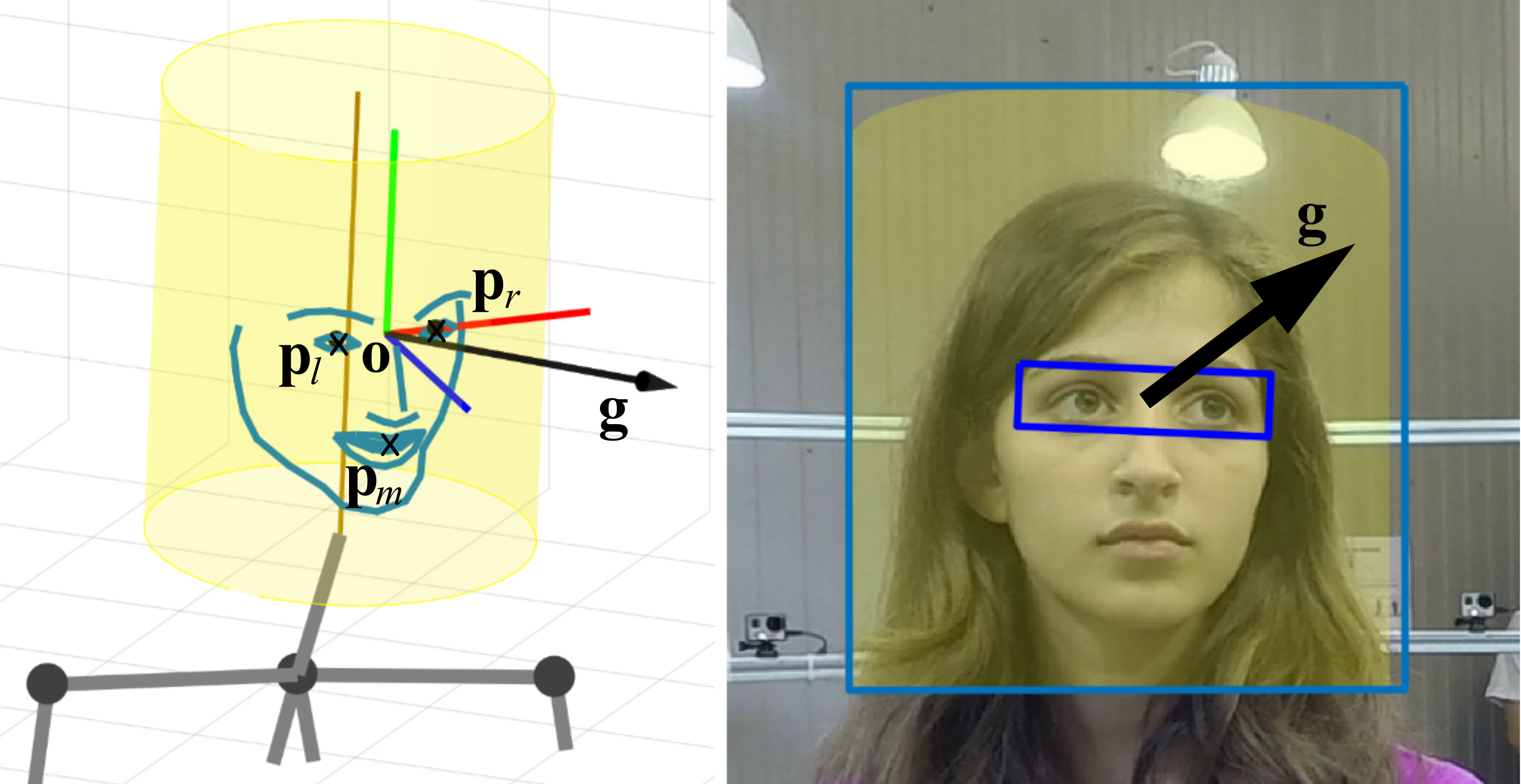}
	\end{center}
    \caption{(Left) 3D geometry of gaze. The black arrow is the gaze direction. The red, green and blue segment are the $x$, $y$, and $z$-axes of the moving head coordinate system. (Right) The gaze geometry is projected onto the image.}
    \label{fig:gaze3}
\end{figure}

\subsection{Gaze}\label{humbi:gaze}
 
HUMBI Gaze contains 93K images (4 gaze directions $\times\sim$30 views per subject). We represent gaze geometry using a unit 3D vector $\mathbf{g}\in\mathds{S}^2$ with respect to the moving head coordinate system. 

The head coordinate is defined as follows. The origin is the center of eyes, $\mathbf{o}=(\mathbf{p}_{l} + \mathbf{p}_{r})/2$, where $\mathbf{p}_{l}, \mathbf{p}_{r}\in\mathbb{R}^3$ are the centers of the left and right eyes (Fig.~\ref{fig:gaze3}). The $x$-axis is the direction along the line joining the two eye centers, $(\mathbf{p}_{l}-\mathbf{o})/\|\mathbf{p}_{l}-\mathbf{o}\|$; the $z$-axis is the direction perpendicular to the plane made of $\mathbf{p}_l$, $\mathbf{p}_{r}$, and $\mathbf{p}_{m}$ where $\mathbf{p}_{m}$ is the center of the mouth, orienting towards the frontal face; the $y$-axis is defined by the cross product between the $z$- and $x$-axes.

\begin{figure}[t]
  \begin{center}
      \includegraphics[height=0.25\textheight]{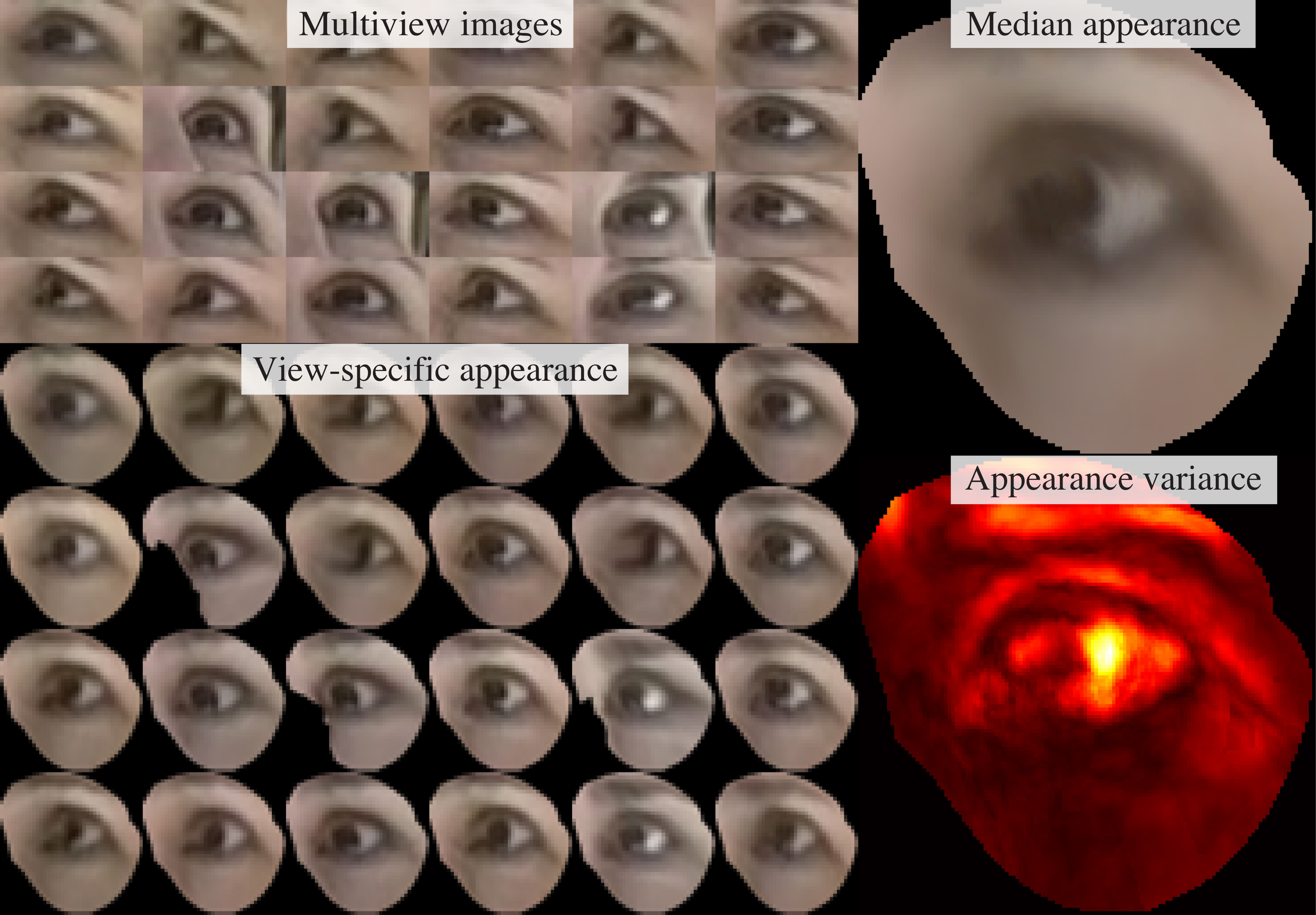}
  \end{center}
  \caption{The view-specific gaze appearance rendered from multiview images with its median and variance. The eye patches from multiview images are normalized to ensure consistent orientation and distance across the views.}
  \label{fig:gaze_uv}
\end{figure}

For eye appearance, we provide two representations: (1) normalized eye patches and (2) pose-independent appearance map. To generate a normalized eye patch, (36 x 60 pixels), we warp an eye patch region to the normalized coordinate such that the orientation and distance remain constant across views. RGB values are histogram-equalized. To construct a pose-independent appearance map, we leverage 3D eye mesh in the Surrey face model~\cite{huber2016multiresolution} to build a canonical atlas coordinate (UV) for each eye. We represent the view-specific appearance map $\mathcal{A}$ (256 $\times$ 256 pixels) by projecting the corresponding pixels in an image onto the atlas coordinate. Fig.~\ref{fig:gaze_uv} illustrates view-specific appearance map across views with median and variance. The variance map indicates that the appearance is highly dependent on viewpoint in particular in the iris region.

\subsection{Face}\label{humbi:face}

HUMBI Face contains 17.3M images (330 frames $\times$ 68 views per subject). We represent face geometry using a 3D blend shape model~\cite{huber2016multiresolution} with 3,448 vertices and 6,736 faces:
\begin{align}
    \mathbf{V}(\boldsymbol{\alpha}^s,\boldsymbol{\alpha}^e) = \mathbf{S}_0 + \sum_{i=1}^{K^s} \alpha_i^s \mathbf{S}_i^s + \sum_{i=1}^{K^e} \alpha_i^e \mathbf{S}_i^e,
\end{align}
where $\mathbf{V}\in \mathbb{R}^{3\times D}$ is the 3D face vertices, $\mathbf{S}_0$ is the mean face, $\mathbf{S}_i^s$ and $\alpha_i^s$ are the $i^{\rm th}$ shape basis and its coefficient, and $\mathbf{S}_i^e$ and $\alpha_i^e$ are the $i^{\rm th}$ expression basis and its coefficient. $D$ is the number of the 3D vertices of the face shape model.

We reconstruct the face model by minimizing the following cost:
\begin{align}
    E = E^{k} + \lambda^{a} E^{a}, \label{eq:face}
\end{align}
where $E^{k}$ and $E^{a}$ are the errors of the 3D keypoint and the appearance, respectively, and $\lambda^{a}=1\times10^{-5}$ is the weight parameter.



We minimize the geometric error between the 3D face model and the reconstructed keypoints:
\begin{align}
E^{k}\big(\mathbf{Q},\boldsymbol{\alpha}^s,\boldsymbol{\alpha}^e) = \sum_i \|\mathcal{K}_i- \mathbf{Q}(\mathbf{V}_i)\|^2 \end{align}
where $\boldsymbol{\alpha}^s\in\mathbb{R}^{63}$ and $\boldsymbol{\alpha}^e\in\mathbb{R}^{6}$ are shape and expression coefficients, $\mathcal{K}_i$ is the $i^{\rm th}$ face keypoint, and $\mathbf{V}_i$ is the pre-defined vertex in $\mathbf{V}$ corresponding to $\mathcal{K}_i$ as approximation. $\mathbf{Q}$ is a global rigid transformation over all vertices in the face mesh.


We model the texture using a linear combination of the texture bases~\cite{bfm09}:
\begin{align}
    \mathbf{T}(\boldsymbol{\alpha}^t) = \mathbf{T}_0 + \sum_{i=1}^{K^t} \alpha_i^t \mathbf{T}_i,
\end{align}
where $\mathbf{T}\in \mathds{R}^{3\times D}$ is the 3D face texture, $\mathbf{T}_0$ is the mean texture model, $\mathbf{T}_i$ and $\alpha_i^t$ are the $i^{\rm th}$ texture basis and its coefficient.

\begin{figure}[t]
	\begin{center}
	    \includegraphics[width=0.45\textwidth]{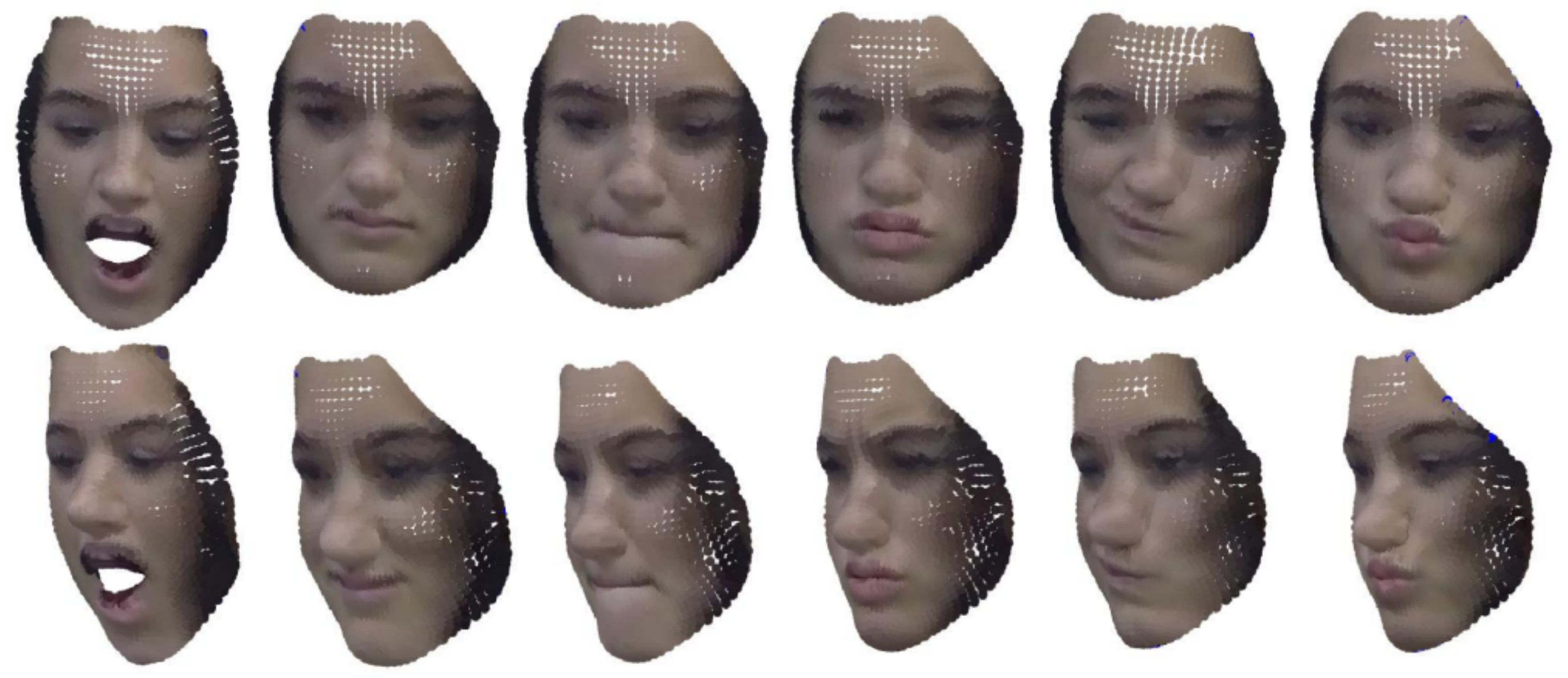}
		\includegraphics[width=0.45\textwidth]{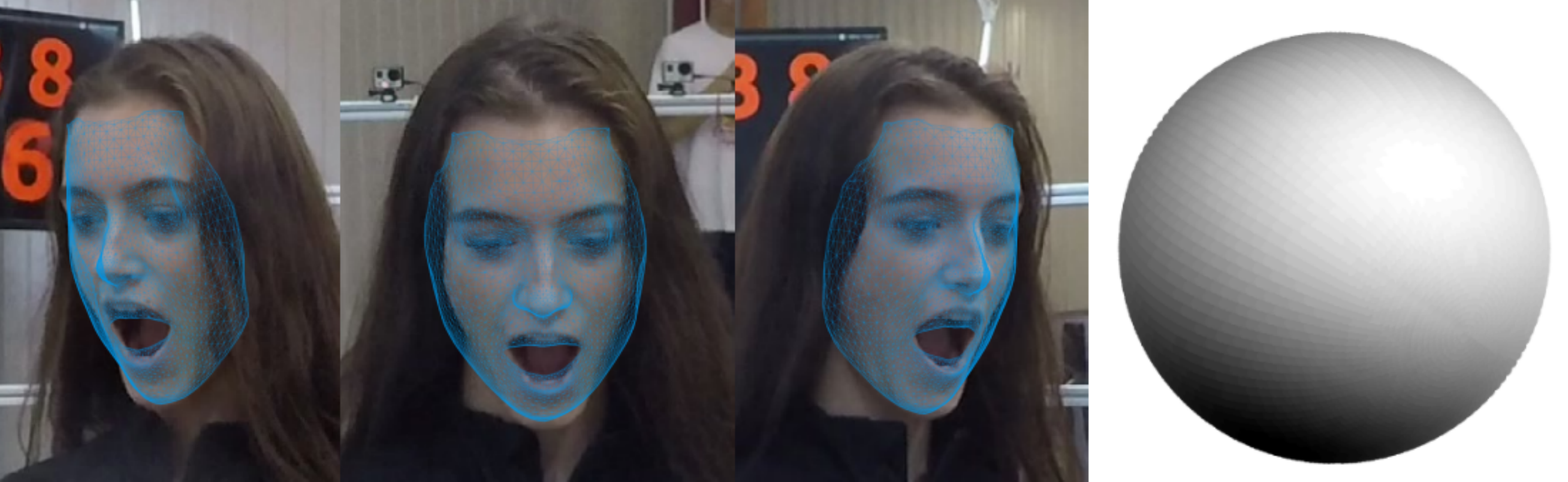}
	\end{center}
	\caption{(Top) The recovered 3D faces with various expressions where the first and second rows describe the rendered face from the front and side views respectively. (Bottom left) Alignment between projected mesh and subject's face. (Bottom right) The estimated illumination using a linear combination of the spherical harmonics.}
    \label{fig:face}
\end{figure}

We render appearance in the $j^{\rm th}$ image using the shape model $\mathbf{V}$, texture $\mathbf{T}$, illumination, and camera projection: $\mathbf{C}_j = g(\mathbf{V}, \mathbf{T}, \boldsymbol{\alpha}^h, \mathbf{P}_j)$ where $\boldsymbol{\alpha}^h$ is the coefficients of the spherical harmonics bases~\cite{arfken1985spherical}, $\mathbf{P}_j\in\mathds{R}^{3\times 4}$ is the projection matrix of the $j^{\rm th}$ camera, and $g$ encapsulates a rendering function based on Lambertian reflectance~\cite{pedrotti2017introduction}. 



The appearance error is defined as:
\begin{align}
    E^{a}(\boldsymbol{\alpha}^s, \boldsymbol{\alpha}^e, \boldsymbol{\alpha}^t, \boldsymbol{\alpha}^h) =  \sum_j \|\mathbf{c}_j - \mathbf{C}_j\|^2,
\end{align}where
$\mathbf{c}_j$ is the face appearance in the $j^{\rm th}$ image.

We optimize Equation~(\ref{eq:face}) using a nonlinear least squares solver~\cite{ceres-solver} with an ambient light initialization. Fig.~\ref{fig:face} illustrates the resulting face reconstruction that includes the shape, expression, texture, and illumination. To learn a consistent shape of the face model for each subject, we infer the maximum likelihood estimate of the shape parameters given the reconstructed keypoints over frames, which allows us to fit to the best model.

\begin{figure}[t]
  \begin{center}
    \includegraphics[height=0.25\textheight]{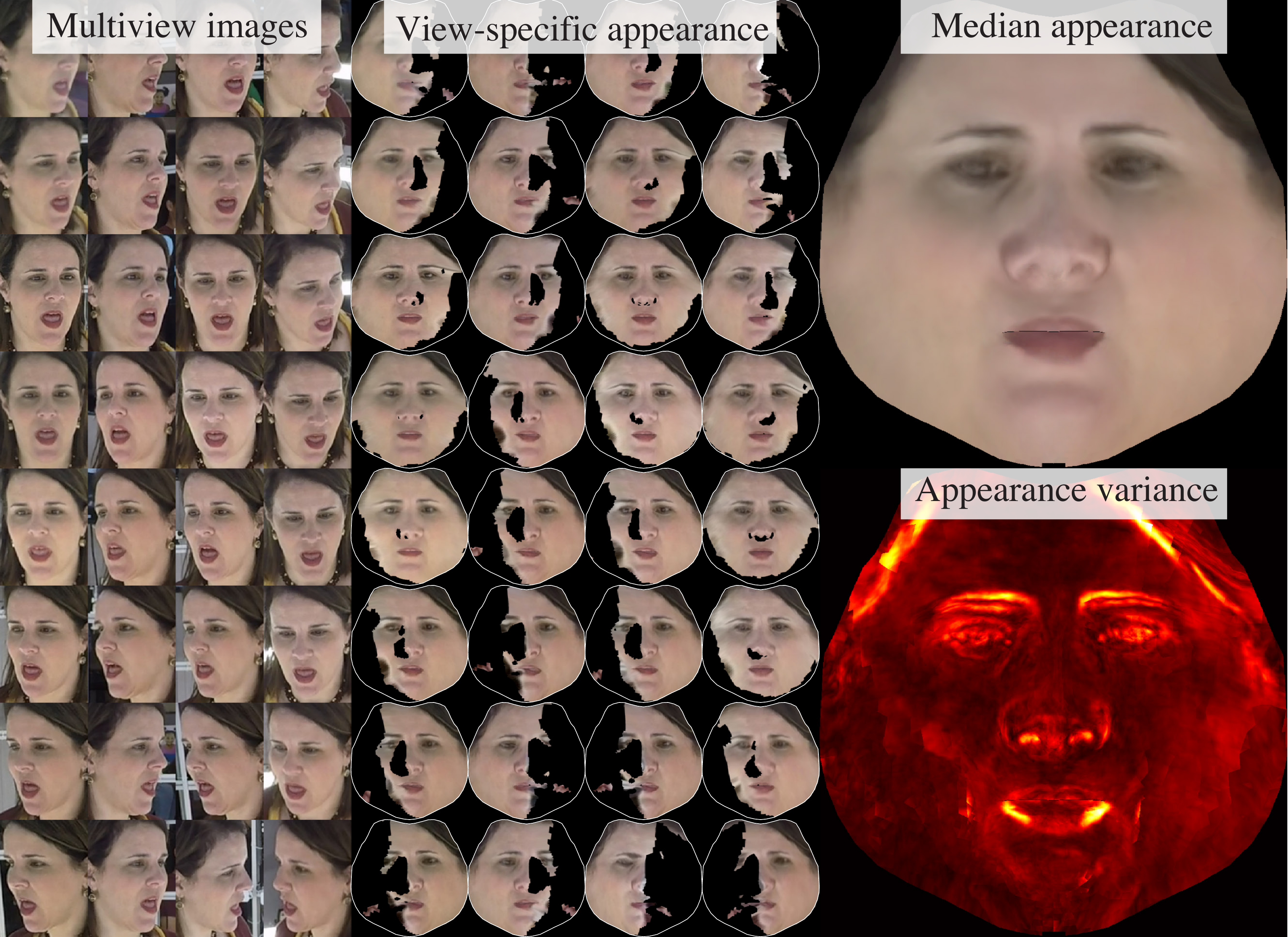}
  \end{center}
  \caption{The view-specific face appearance rendered from multiview images with its median and variance. The mesh parameterization of the fitted 3D face model~\cite{huber2016multiresolution} is used to project the images to UV coordinates.}
  \label{fig:face_uv}
\end{figure}

Given the reconstructed face model, we construct a view-specific appearance map $\mathcal{A}_{\rm face}$ (1024 x 1024 pixels) by projecting the pixels in an image onto its canonical atlas coordinate. For each UV coordinate in the appearance map, we find the corresponding coordinate in the image via 3D mesh with bilinear interpolation. Fig.~\ref{fig:face_uv} illustrates the view-specific appearance across views with its median and variance. The variance map shows that the appearance is dependent on views, e.g. the regions of salient landmarks such as eye, eyebrows, nose, and mouth, which justifies the necessity of view-specific appearance modeling~\cite{LOMBARDI:2018}.

\begin{figure}[h]
  \begin{center}
      \includegraphics[height=0.227\textheight]{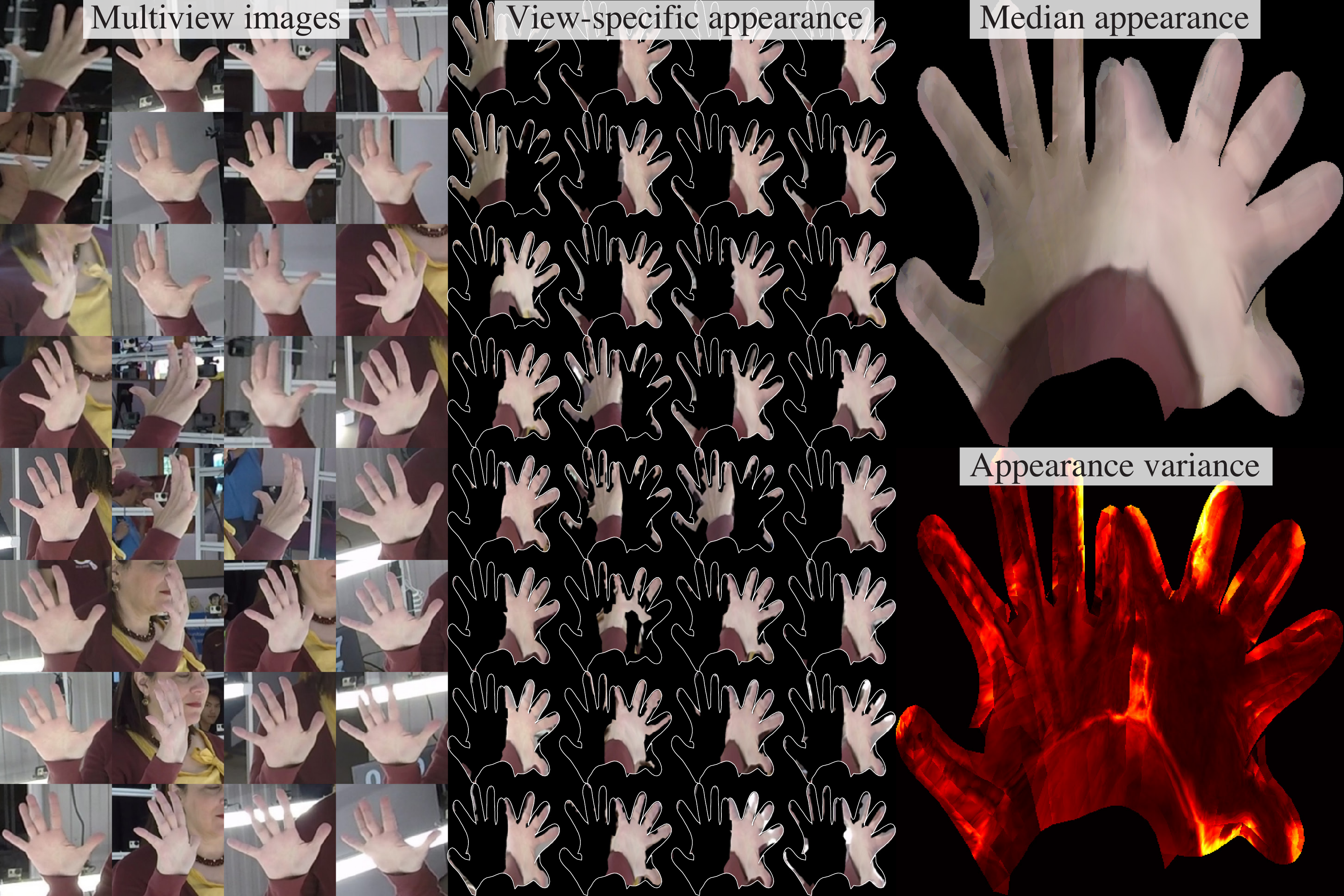}
  \end{center}
  \caption{The view-specific hand appearance rendered from multiview images with its median and variance. The mesh parameterization of the fitted 3D hand model~\cite{MANO:SIGGRAPHASIA:2017} is used to project the images to UV coordinates.}
  \label{fig:hand_uv}
\end{figure}

\subsection{Hand}\label{humbi:hand}
HUMBI Hand contains 24M images (290 frames $\times$ 68 views per subject). We represent hand geometry using a 3D parametric model~\cite{MANO:SIGGRAPHASIA:2017} with 778 vertices and 1,538 faces:
\begin{align}
    \mathbf{V}(\boldsymbol{\alpha}^\theta,\boldsymbol{\alpha}^\beta) = \mathbf{S}_0 + \sum_{i=1}^{K^\theta} \alpha_i^\theta \mathbf{S}_i^\theta + \sum_{i=1}^{K^\beta} \alpha_i^\beta \mathbf{S}_i^\beta,
\end{align}
where $\mathbf{V}\in \mathbb{R}^{3D}$ is the 3D hand vertices, $\mathbf{S}_0$ is the mean hand, $\mathbf{S}_i^\theta$ and $\alpha_i^\theta$ are the $i^{\rm th}$ pose basis and its coefficient, and $\mathbf{S}_i^\beta$ and $\alpha_i^\beta$ are the $i^{\rm th}$ shape basis and its coefficient. $D$ is the number of the vertices of the hand shape model.

We reconstruct the 21 hand keypoints $\mathcal{K}$ by minimizing the following cost:
\begin{align}
E = E^k + \lambda^\theta E^\theta + \lambda^\beta E^\beta,
\end{align}
where $\lambda^\theta=3\times10^{-3}$ and $\lambda^\beta=1\times10^{-2}$ are the weights for the pose and shape regularization, respectively.

Given the correspondences between the reconstructed keypoints and the hand mesh vertices, the keypoint error is defined as:
\begin{align}
    E^k(\mathbf{Q}, \boldsymbol{\alpha}^\theta, \boldsymbol{\alpha}^\beta) = \sum_i \|\mathcal{K}_i - \mathbf{Q} (\mathbf{V}_i)\|^2,
\end{align}
where $\mathbf{V}_i$ is the vertex in $\mathbf{V}$ corresponding to the $i^{\rm th}$ hand keypoint and $\mathbf{Q}$ is a rigid transformation between the reconstructed keypoints and the corresponding hand mesh vertices.

We apply regularization on the pose and shape parameters:
\begin{align}
    E^\theta (\boldsymbol{\alpha}^\theta) = \|\boldsymbol{\alpha}^\theta\|^2, \qquad E^\beta (\boldsymbol{\alpha}^\beta) = \|\boldsymbol{\alpha}^\beta\|^2.
\end{align}
We compute an initial estimate of the rigid transformation using the 6 keypoints in the palm, and estimate the shape and pose parameters in an alternating fashion followed by a nonlinear optimization. For each subject, we estimate common hand shape parameters as the median of shapes over time.

Given the reconstructed hand mesh model, we construct a view-specific appearance map $\mathcal{A}$ (512 $\times$ 512 pixels) by projecting the pixels in an image onto the canonical atlas coordinate. Fig.~\ref{fig:hand_uv} illustrates the view-specific appearance across views with its median and variance of appearance. The variance map shows that the appearance is dependent on viewpoints.



\begin{figure}[t]
	\begin{center}
		\includegraphics[width=0.45\textwidth]{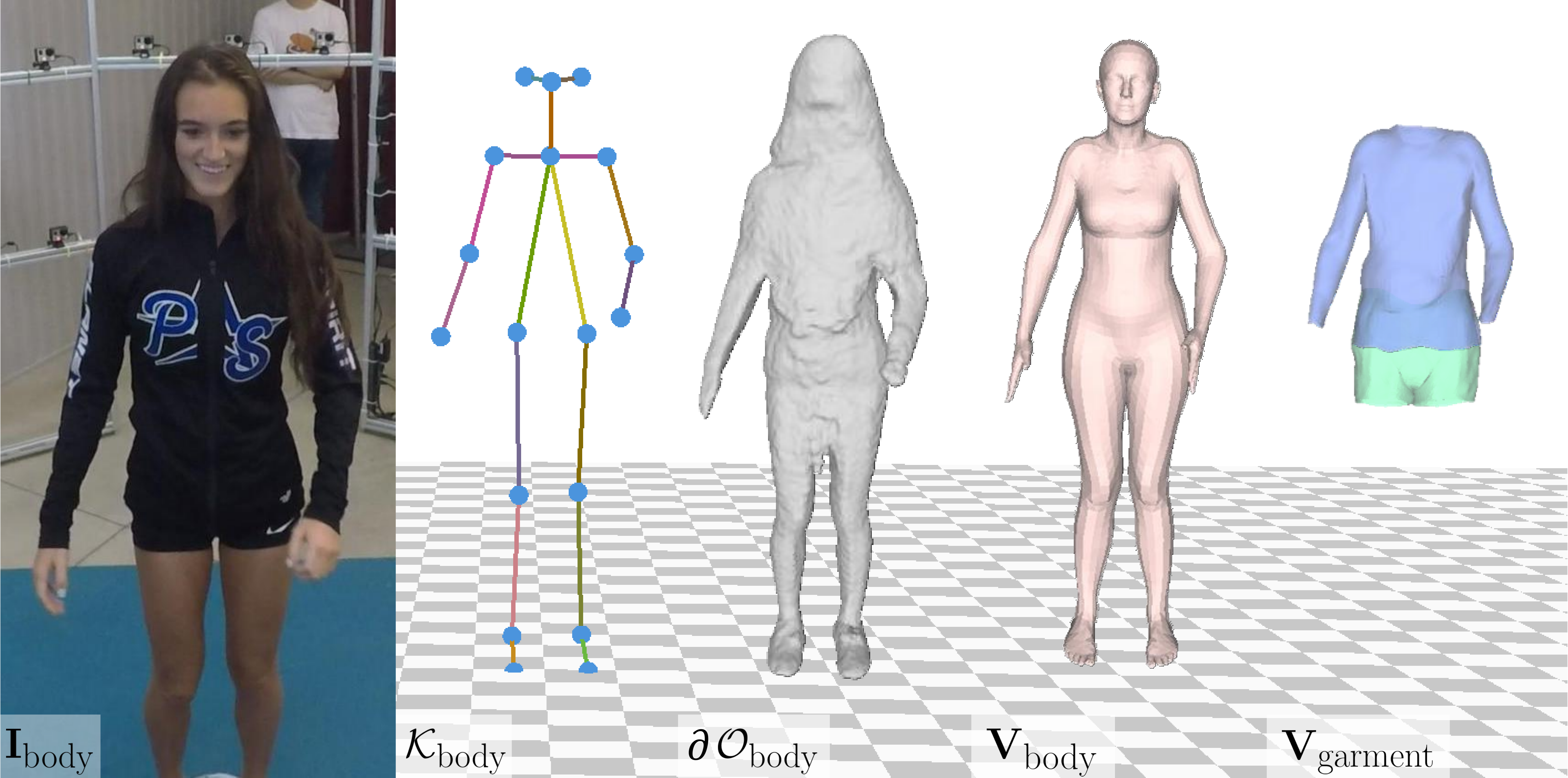}
	\end{center}
    \caption{Results for body $\mathbf{V}_{\rm body}$ and clothing $\mathbf{V}_{\rm garment}$ model fitting where the 3D reconstruction of the keypoints $\mathcal{K}_{\rm body}$ and the surface $\partial{\mathcal{O}}_{\rm body}$ are used as spatial priors.}
    \label{fig:body_vis}
\end{figure}

\subsection{Body}\label{humbi:body}
HUMBI Body contains 26M images (315 frames $\times$ 107 views per subject). We present body geometry using a 3D linear blend shape model~\cite{loper2015smpl} with 4,129 vertices and 7,999 faces without hand and head parts:
\begin{align}
    \mathbf{V}(\boldsymbol{\beta},\boldsymbol{\theta}) = {W}(\bar{\mathbf{V}}+\mathcal{T}(\boldsymbol{\beta},\boldsymbol{\theta}),\mathcal{K}(\boldsymbol{\beta}),\boldsymbol{\theta},\mathcal{W}),
\end{align}
where $\mathbf{V}\in \mathbb{R}^{3\times D}$ is the vertices of the posed 3D body ($D=4,129$), and $W$ is the skinning function~\cite{loper2015smpl} that takes the mean body shape in the rest pose $\bar{\mathbf{V}}$, pose and shape blending shapes $\mathcal{T}$, 3D keypoints $\mathcal{K}$, and blending weights $\mathcal{W}$. This skinning function is parameterized by the shape $\boldsymbol{\beta}\in\mathds{R}^{10}$ and pose coefficients $\boldsymbol{\theta}\in\mathds{R}^{24\times3}$ with axis-angle representation, where ${\boldsymbol{\theta}}_1\in \mathbb{R}^{1\times 3}$ represents the root orientation and others the relative angles with respect to the root joint.

We reconstruct the body model by minimizing the following cost:
\begin{align}
E(\boldsymbol{\theta},\boldsymbol{\beta}, \boldsymbol{t},\boldsymbol{s})= E^{p}+ \lambda^{s}E^{s}+ \lambda^{r}E^{r},
\label{eq:optim_body}
\end{align}
where $\lambda^{s}=5\times10^{-3}$ and $\lambda^{r}=1\times10^{-5}$ control the importance of each measurement, and $\boldsymbol{t}\in\mathds{R}^{3}$ and $\boldsymbol{s}\in\mathds{R}^{+}$ represent the global translation and scale, respectively.  



Given the correspondences between the reconstructed keypoints (i.e., $\mathcal{K}_{\rm body}$ in Fig.~\ref{fig:body_vis}) and the body mesh, we recover the posed body model by minimizing the keypoints error:
\begin{align}
E^{p}(\boldsymbol{\theta},\ \boldsymbol{\beta},\boldsymbol{t},\boldsymbol{s})=\sum_i \left\|\mathcal{K}_{i} - {\mathbf{V}}_{i}\right\|^2.
\label{eq:data}
\end{align}



We recover the shape of the body model by aligning the model to the surface of the 3D reconstruction (i.e., $\partial\mathcal{O}_{\rm body}$ in Fig.~\ref{fig:body_vis}). We use Chamfer distance to measure their alignment:
\begin{align}
 E^{s}(\boldsymbol{\beta},\ \boldsymbol{\theta},\ \boldsymbol{t},\boldsymbol{s}) =  d_{\rm chamfer}(\partial\mathcal{O},\mathbf{V}),
\label{eq:icp2}
\end{align}
where $d_{\rm chamfer}$ measures Chamfer distance between two sets of point clouds, $\partial{\mathcal{O}}\in\mathds{R}^{3\times N}$ is a set of the 3D points on the outer surface of the occupancy map, and $N$ is the number of the 3D points. We use Shape-from-silhouette\footnote{MultiView stereo reconstruction~\cite{schoenberger2016sfm} is complementary to the shape-from-silhouette.}~\cite{Laurentini94} to reconstruct the occupancy map $\mathcal{O}$ with human body parts segmentation~\cite{lin2017refinenet}. 



$E^r$ regularizes the estimated shape by minimizing the difference of the estimated shape $\boldsymbol{\beta}^{t}$ and the ones from the previous time instances, which helps to prevent a degenerate reconstruction due to the measurement noise/error, e.g., erroneous surface reconstruction of the body parts due to the occlusion by hands:
 \begin{align}
 E^{r}(\boldsymbol{\beta}^{t};\boldsymbol{\beta}^{T})= \left\|\boldsymbol{\beta}^{t} - {\rm median}(\boldsymbol{\beta}^{T})\right\|^2,
\label{eq:prior}
\end{align}
where $\boldsymbol{\beta}^{\rm T}=\{\boldsymbol{\beta}^{1},\boldsymbol{\beta}^{2}, ...,\boldsymbol{\beta}^{t-1}\}$ is the set of shape parameters estimated from previous time instances, and $\rm median(\cdot)$ is the median operator. For $t=1$, only L2 regularization is applied, i.e., $E^{r}(\boldsymbol{\beta}^{1})= \left\|\boldsymbol{\beta}^{t}\right\|^{2}$.

\begin{figure}[t]
  \begin{center}
    \includegraphics[height=0.227\textheight]{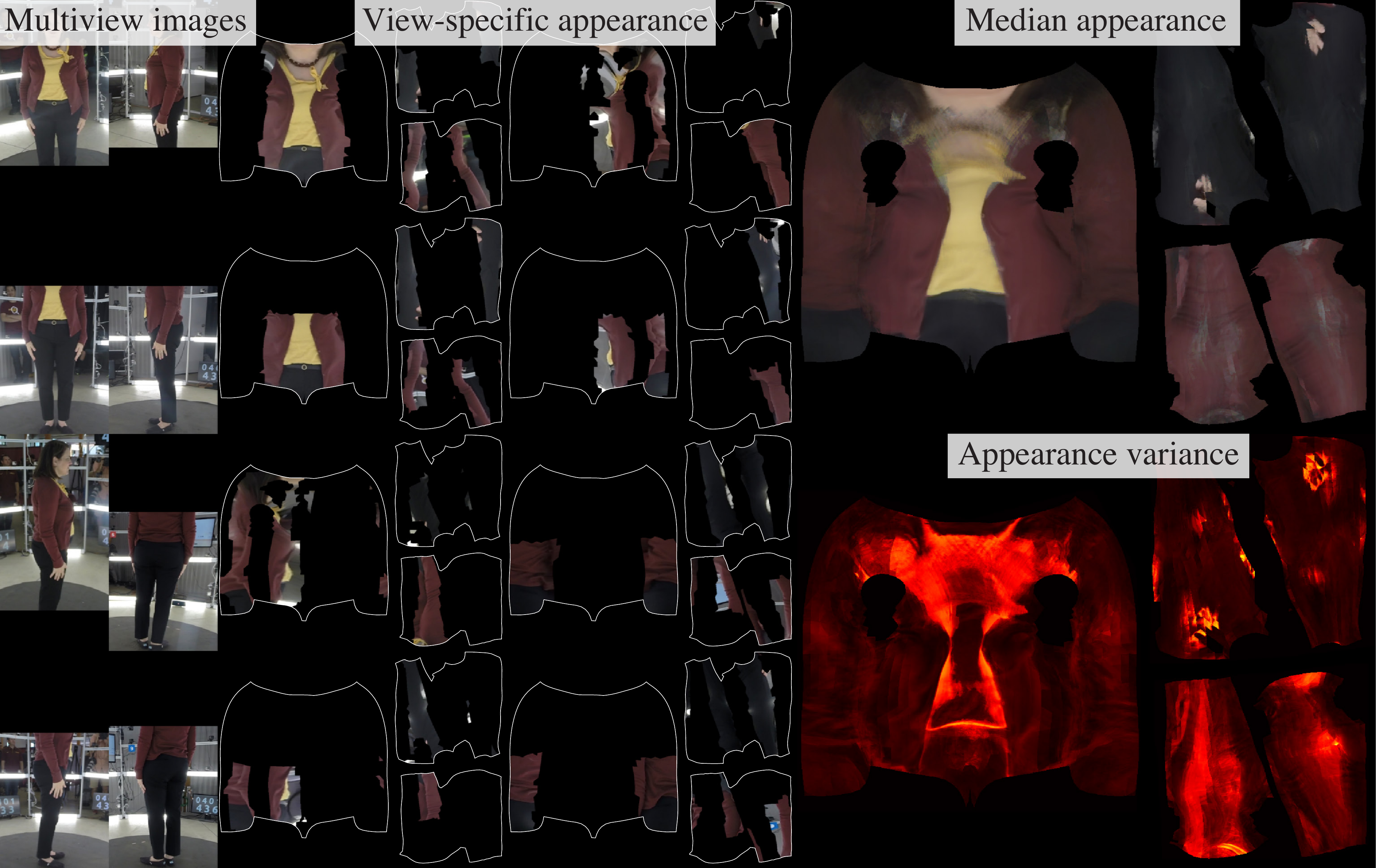}
  \end{center}
  \caption{The view-specific body appearance rendered from multiview images with its median and variance. The mesh parameterization of the fitted 3D body model~\cite{loper2015smpl} is used to project the images to UV coordinates.}
  \label{fig:body_uv}   
\end{figure}

Given the reconstructed body mesh model, we construct a view-specific appearance map $\mathcal{A}$ (1024 $\times$ 1024 pixels) by projecting the pixels in an image onto the canonical atlas coordinate. Fig.~\ref{fig:body_uv} illustrates view-specific appearance across views with its median and variance of appearance. The variance map shows that the appearance is dependent on viewpoints.

{
\renewcommand{\tabcolsep}{4pt} 
\begin{table}[t]
\setlength{\extrarowheight}{2pt}
\centering
\scriptsize
\begin{tabular}{l||c|c|c|c}
\hline
Bias / Variance     &UTMV~\cite{sugano:2014} &MPII~\cite{zhang15_cvpr} &RT-GENE~\cite{fischer2018rt} &HUMBI\\
\hline
Gaze  &\textbf{7.43} / \textbf{33.09} &8.80 / 10.10 &19.35 / \underline{31.71} &\underline{7.70} / 30.01\\
\hline
Head pose &\underline{4.20} / \textbf{29.28} &12.51 / 16.04 &17.97 / 22.48 &\textbf{1.42} / \underline{24.77}\\
\hline
Eye pose &\underline{8.43} / 15.40 &20.81 / \underline{19.02} &\textbf{3.21} / 17.49 &8.78 / \textbf{19.04}\\
\hline
Average &\underline{6.69} / \textbf{25.93} &14.04 / 15.05 &13.51 / 23.90 &\textbf{5.98} / \underline{24.61}\\
\hline
\end{tabular}
\caption{Bias and variance analysis of the distribution of head pose, gaze and eye pose (unit: degree, smallest bias and largest variance in bold, second with underline).}
\label{table:gaze_distribution_table}
\end{table}
}
\begin{figure}[t]
	\vspace{-3mm}
	\begin{center}
		\includegraphics[width=3.35in]{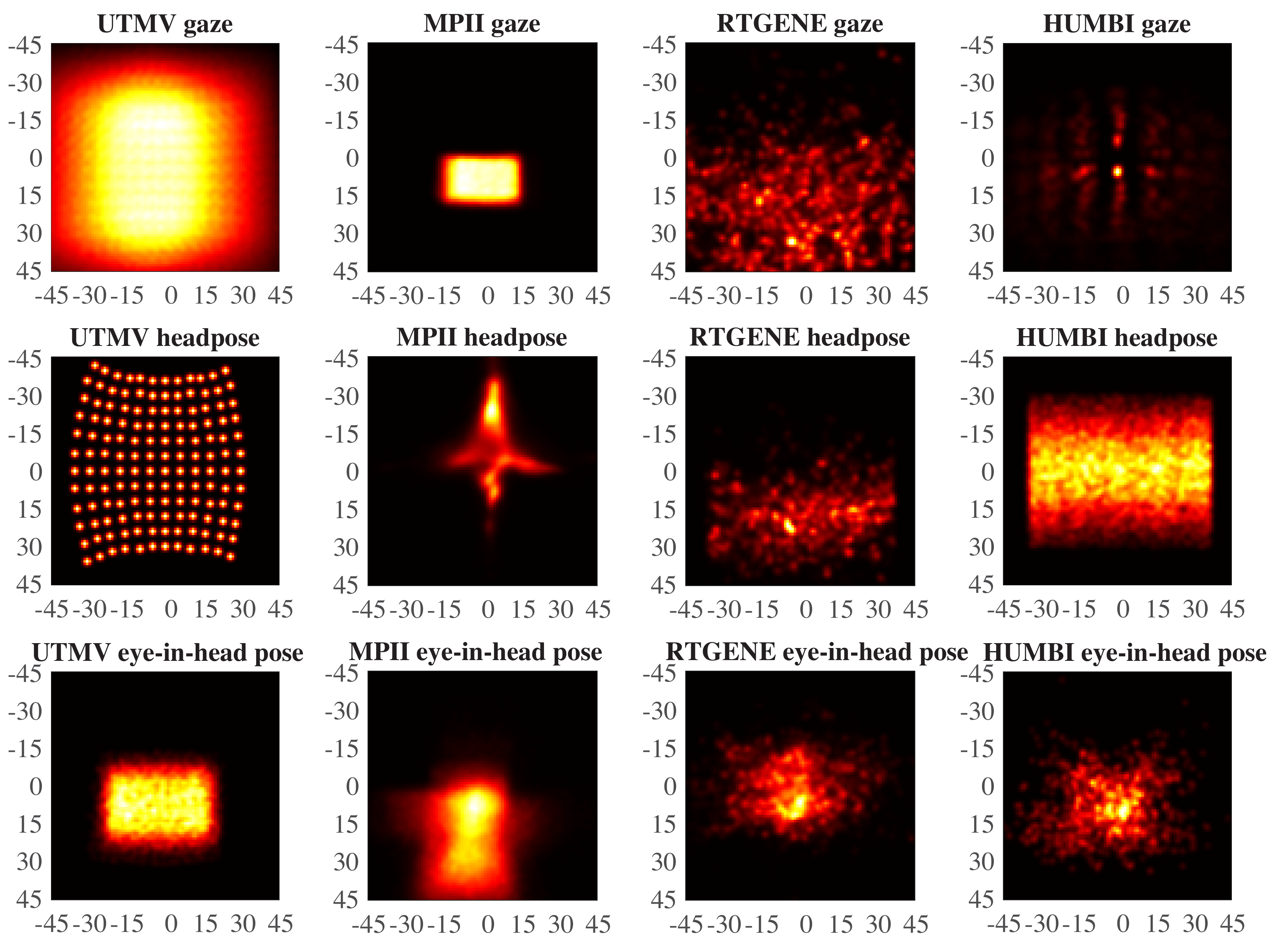}
	\end{center}
	\vspace{-3mm}
    \caption{The distributions of the head pose, gaze, and eye-in-head poses for RT-GENE~\cite{fischer2018rt}, MPII-Gaze~\cite{zhang15_cvpr}, UT-Multiview~\cite{sugano:2014}, and HUMBI. Horizontal and vertical axis represent yaw and pitch angles, respectively (unit: degree). RT-GENE captures 15 subjects using a mobile eye-tracking device which covers a wide range of gaze movements with limited headposes. MPII-Gaze captures 15 subjects using a single webcam on a laptop where the gaze and head poses are highly biased, resulting in the small range of gaze distribution. UTMV captures 50 subjects with fixed head pose using 8 cameras mounted on a monitor. Note that UTMV is composed of the synthesized eye images from dense virtual viewpoints. For HUMBI, 772 subjects are captured from 30 real viewpoints where a subject can move head freely. HUMBI spans a wider and more continuous range of head poses than other datasets with comparable gaze and eye-in-head pose distribution.}
    \label{fig:gaze_distribution}
\end{figure}

\subsection{Garment}\label{humbi:cloth}
Similar to HUMBI Body, HUMBI Garment includes 26M images. Given the body reconstruction, we represent the garment geometry using an in-house garment mesh model\footnote{
A similar approach was used to reconstruct garment from 4D scanner~\cite{pons2017clothcap}.} $\mathbf{V}\in\mathds{R}^{3\times D}$, where $D$ is the number of 3D points. Unlike parametric models used in face, hand, and body, there exists no shape model that can express diverse topology, style, and type of garments. Instead, we represent the dynamic garment shape with per-vertex 3D warping fields~\cite{newcombe2015dynamicfusion} that map the garment mesh vertices at the rest pose $\bar{\mathbf{V}}$ to the deformed garment:
\begin{align}
    \mathbf{V}_i = \mathbf{R}_{i}\bar{\mathbf{V}}_i + \mathbf{t}_{i},
\end{align}
where $(\mathbf{R}_i, \textbf{t}_i)\in SE(3)$ is the 6\textit{D} transformation. We optimize this warping field by minimizing the following objective:
\begin{align}
E(\mathbf{R},\mathbf{t})=E^{c}+\lambda^{o}E^{o}+\lambda^{g}E^{g},
\label{eq:optim}
\end{align}
where $\lambda^{o}=1\times10^{-1}$ and $\lambda^{g}=1\times10^{-1}$ are weight parameters.

We predefine a set of fiducial correspondences between the garment and body meshes, which are the control points to deform the garment mesh. $E^{c}$ measures the correspondence error:
\begin{align}
E^{c}(\mathbf{R},\mathbf{t})=\sum_{i}\|\widehat{\mathbf{V}}^{b}_{i}-\widehat{\mathbf{V}}^{g}_{i}\|^2,
\end{align}
where $\widehat{\mathbf{V}}^{g}$ and $\widehat{\mathbf{V}}^{b}$ are the corresponding vertices between the garment and body model, respectively.

$E^{o}$ measures Chamfer distance to align the garment mesh model with the 3D points on the outer surface of the occupancy map $\partial\mathcal{O}\in\mathds{R}^{3\times N}$ where $N$ is the number of the correspondences: 
\begin{align}
    E^{o}(\mathbf{R},\mathbf{t}) =  d_{\rm chamfer}(\partial{\mathcal{O}},\mathbf{V}).
\end{align}

$E^{g}$ is a spatial regularization based on Laplacian mesh deformation~\cite{sorkine2007rigid} that enforces as-rigid-as-possible deformation by penalizing a non-smooth and non-rigid vertex with respect to its neighboring vertices: 
\begin{align}
    E^{g}(\mathbf{R},\mathbf{t}) = \nabla^2 \mathbf{V}.
\end{align}

We use three garment topologies, i.e., tops: sleeveless shirts (3,763 vertices and 7,261 faces), T-shirts  (6,533 vertices, 13,074 faces), and long-sleeve shirts (8,269 vertices and 16,374 faces), and bottoms: short (3,975 vertices and 7,842 faces), medium (5,872 vertices and 11,618 faces), and long pants (11,238 vertices and 22,342 meshes), which are manually matched to each subject.

\vspace{-1mm}
\section{Evaluation}
\vspace{-1mm}
We evaluate HUMBI in terms of generalizability, diversity, and accuracy. For generalizability, we conduct the cross-data evaluation on the tasks of single view human reconstruction, e.g., monocular 3D face mesh prediction. For diversity, we characterize the distribution of HUMBI to analyze the valid range of prediction, e.g., gaze direction distribution along the yaw and pitch angle. For accuracy, we measure the silhouette similarity between the human annotation and the reprojection of the reconstructed 3D model.
{
\renewcommand{\tabcolsep}{3pt} 
\begin{table}[t]
\setlength{\extrarowheight}{2pt}
\centering
\scriptsize
\begin{tabular}{l||c|c|c|c|c|c}
\hline
\multirow{2}{*}{\backslashbox[20mm]{Testing}{Training}} & \multirow{2}{*}{MPII} & \multirow{2}{*}{UTMV} & \multirow{2}{*}{HUMBI} & MPII & UTMV & All \\
& & & & + HUMBI & + HUMBI & \\
\hline
MPII~\cite{zhang15_cvpr}  & 6.1$\pm$3.3 &11.8$\pm$6.6  &8.8$\pm$4.8 & 7.4$\pm$4.1 &7.7$\pm$4.6 &7.5$\pm$4.3\\
\hline
UTMV~\cite{sugano:2014}  & 23.3$\pm$9.4 & 5.0$\pm$3.2 & 8.2$\pm$4.5 &9.4$\pm$5.1 &5.4$\pm$3.2 &6.3$\pm$3.8\\
\hline
HUMBI & 23.7$\pm$13.7 & 14.6$\pm$10.3 & 7.9$\pm$5.4 & 8.9$\pm$6.2 &8.0$\pm$5.4 &8.3$\pm$5.5\\
\hline
\end{tabular}
\caption{The mean error of 3D gaze prediction for the cross-data evaluation (unit: degree). Training with HUMBI shows the minimum performance degradation across the datasets, implying the strong generalization ability.}
\label{table:gaze_experiment}
\end{table}
}
\subsection{Gaze}
\noindent\textbf{Baseline Datasets} We use three baseline datasets: (1) MPII-Gaze (MPII)~\cite{zhang15_cvpr} contains 213,659 images from 15 subjects, which was captured under the scenarios of everyday laptop use. (2) UT-Multiview (UTMV)~\cite{sugano:2014} is composed of 50 subjects with 160 gaze directions captured by 8 monitor-mounted cameras. Using the real data, the synthesized images from 144 virtual cameras are augmented. (3) RT-GENE~\cite{fischer2018rt} contains 122,531 images of 15 subjects captured by eye-tracking glasses.

\noindent\textbf{Distribution of Gaze Directions} To characterize HUMBI Gaze, we visualize three measures in Fig.~\ref{fig:gaze_distribution}: (1) gaze pose: the gaze direction with respect to camera pose; (2) head pose: the head orientation with respect to the camera pose; and (3) eye-in-head pose: the gaze direction with respect to the head. HUMBI Gaze spans a wide and continuous range of head poses due to numerous views and natural head movements by many subjects. The yaw and pitch angles of the gaze and eye-in-head poses are distributed uniformly. The quantitative analysis of the bias and variance of the gaze distribution based on Fig.~\ref{fig:gaze_distribution} is summarized in Table~\ref{table:gaze_distribution_table}. HUMBI shows the smallest average bias (5.98$^\circ$ compared with 6.69$^\circ\sim$14.04$^\circ$ from other datasets) and second-largest average variance (24.61$^\circ$ compared with 25.93$^\circ$ of UTMV). While combining HUMBI with other datasets does not improve the performance of the model trained solely on each dataset due to large domain gaps across datasets (e.g., UTMV is synthetic while HUMBI is real as described in Fig.~\ref{fig:gaze_distribution}), it indeed mitigates the domain gap between the data distribution. For example, the models trained on MPII and MPII+HUMBI exhibit 17.2$^\circ$ and 2$^\circ$ of performance drops, respectively, when tested on UTMV. This implies that HUMBI is complementary to other datasets for domain generalization.
%


\begin{figure}[t]
	\begin{center}
	\includegraphics[clip,width=3.6in]{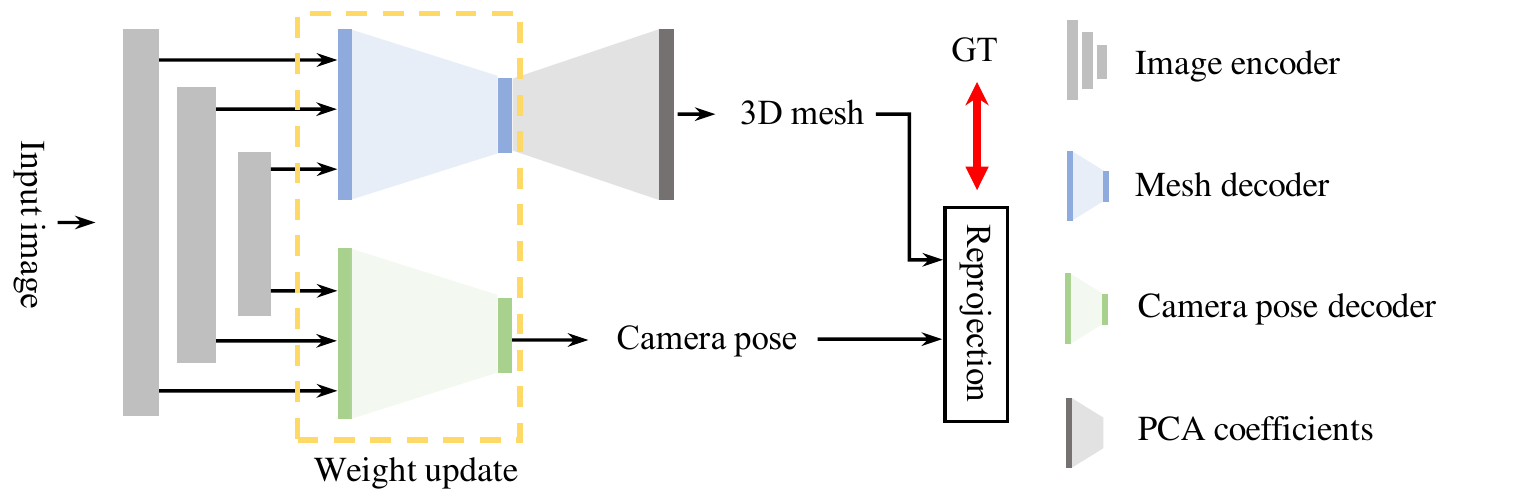}
	\end{center}
\vspace{-4mm}
    \caption{{We use a vanilla network~\cite{Yoon_2019_CVPR} design to evaluate the strength of the datasets. This network takes as input an image and outputs the parameters of the 3D mesh (face, hand, and body) and camera pose. The network is made of the pre-trained image encoder~\cite{simonyan2014very} that extracts image features and two decoders that predict the latent mesh parameters and camera pose where we train these decoders from scratch by minimizing the reprojection error. From the predicted model parameters, we reconstruct the 3D body shape using the PCA coefficients of each body model, i.e., 3DMM~\cite{bfm09} for face, SMPL~\cite{loper2015smpl} for body, and MANO~\cite{MANO:SIGGRAPHASIA:2017} for hand. For cross-dataset evaluation, the network design is fixed while the type of dataset is switched.
}}
    \label{fig:network2}
\end{figure}

\begin{table}[t]
\vspace{3mm}
\setlength{\extrarowheight}{2pt}
\centering
\footnotesize
\begin{tabular}{l|c|c|c}
\hline
\backslashbox[20mm]{Testing}{Training}    &  3DDFA & HUMBI & 3DDFA+HUMBI \\
\hline
3DDFA~\cite{zhu2016face} & 7.1$\pm$6.4 & 20.7$\pm$7.1  &\textbf{4.3$\pm$6.6} \\
\hline
HUMBI & 23.5$\pm$13.9 & 13.3$\pm$13.7 & \textbf{8.4$\pm$12.2}  \\
\hline
\end{tabular}
\caption{The mean error of 3D face mesh prediction for cross-data evaluation (unit: pixel). HUMBI is complementary to existing dataset where training with the combined dataset brings out notable improvement in prediction accuracy.}
\label{table:face_experiment}
\end{table}

\noindent\textbf{Monocular 3D Gaze Prediction}
To validate the generalizability of HUMBI Gaze, we use an existing gaze detection network~\cite{zhang15_cvpr} to conduct a cross-data evaluation. We randomly choose 25K images (equally distributed among subjects) as the experiment set for each dataset. One dataset is used for training, and others are used for testing. A data sample is defined as $\{(\mathbf{e},\mathbf{h}), \mathbf{g}\}$, where $\mathbf{e}\in\mathbb{R}^{36\times60}$, $\mathbf{h}\in\mathbb{R}^{2}$ and $\mathbf{g}\in\mathbb{R}^{2}$ are the normalized eye patch (Sec.~\ref{humbi:gaze}), yaw and pitch angles of the head pose, and the gaze direction with respect to corresponding virtual camera. The network takes as input $\mathbf{e}$ and $\mathbf{h}$ and outputs $\mathbf{g}$.
The detection network is trained to minimize the mean squared error of the gaze yaw and pitch angles. We evaluate each dataset with 90\%/10\% of training/testing split. Table~\ref{table:gaze_experiment} summarize the experiment results. The detector trained by MPII and UTMV shows inferior performance on cross-data evaluation comparing to HUMBI with 3$^\circ$-16$^\circ$ margin. HUMBI exhibits strong performance on cross-data evaluation with minimal degradation (less than 1$^\circ$ drop). Also, UTMV + HUMBI and MPII + HUMBI outperform each alone by a margin of 4.1$^\circ$ and 13.9$^\circ$ when tested on the third dataset MPII and UTMV respectively, showing that HUMBI is highly complementary to UTMV and MPII.

    \begin{figure}[t]
	\begin{center}
\hspace{-4mm}\includegraphics[width=0.4\textwidth]{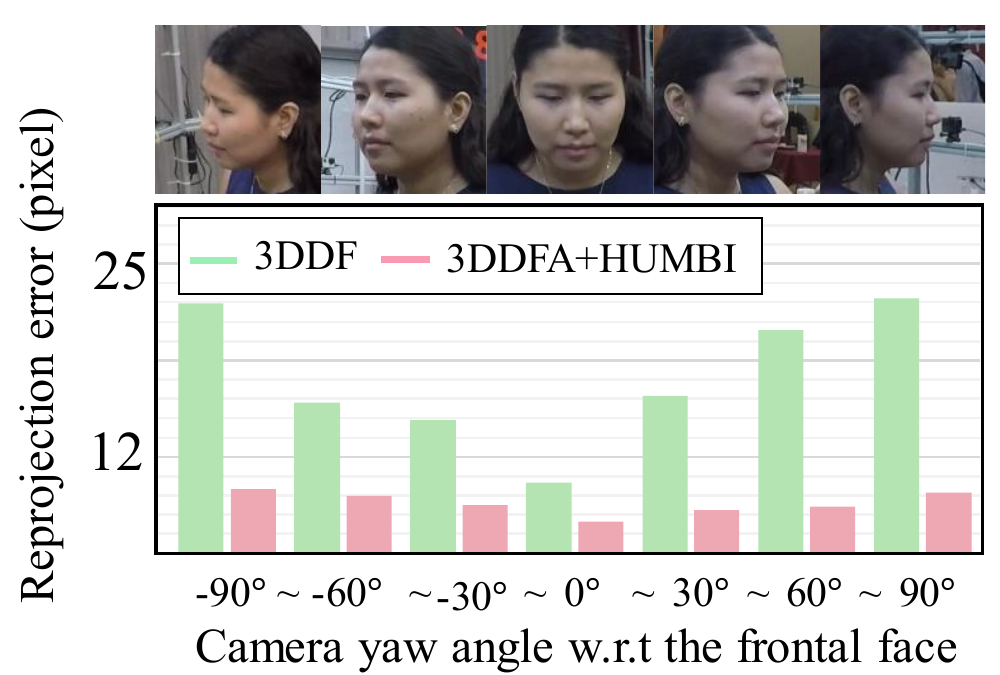}
	\end{center}
	\vspace{-3mm}
    \caption{We evaluate the viewpoint dependency of the face mesh reconstruction models. Combining with HUMBI enforces learning a representation agnostic to viewpoints. $0^{\circ}$ and $\pm 90^{\circ}$ represent the camera angle of the front and side views, respectively.}
    \label{fig:view_aug_face}
\end{figure}

\begin{figure}[t]
	\begin{center}
		\includegraphics[width=0.45\textwidth]{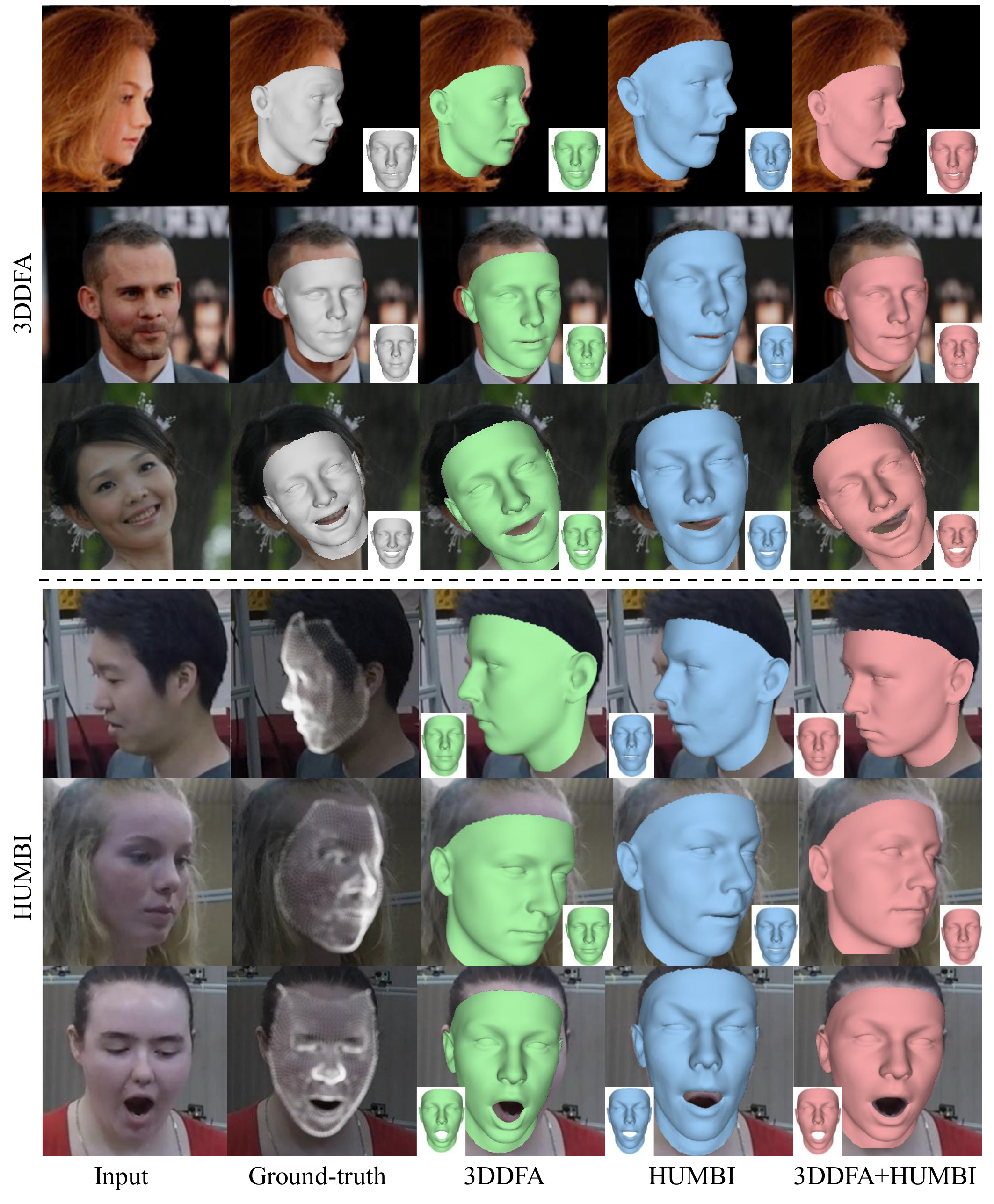}
	\end{center}
	\vspace{-5mm}
    \caption{ The qualitative results of the monocular 3D face prediction network trained with different dataset combination. The column and row represent the type of training and testing data, respectively. Note that, training with the combined datasets improves the prediction of headpose and facial expression.}
    \label{fig:face_res}
\end{figure}



\subsection{Face}\label{exp:face}
\noindent\textbf{Baseline Dataset} We use 3DDFA~\cite{zhu2016face} that provides 6K 2D-3D pairs of the 3D face geometry and the associated images. We use 90\%/10\% of training/testing split. The base face model of 3DDFA is the Basel model~\cite{bfm09}, which is different from our face model (Surrey~\cite{huber2016multiresolution}). We manually establish the correspondences between the two models.

\noindent\textbf{Monocular 3D Face Mesh Prediction}\label{exp:monoface} 
We evaluate HUMBI Face by predicting a 3D face mesh using a vanilla mesh reconstruction network~\cite{Yoon_2019_CVPR} as described in Fig.~\ref{fig:network2}. The network is composed of an image encoder that extracts image features and two decoders: mesh decoder that predicts the shape coefficients and camera pose decoder that estimates the camera extrinsic parameters. We use the pre-trained image encoder~\cite{Yoon_2019_CVPR} while two decoders are trained from scratch.
We train the network using three combinations of datasets, i.e., 3DDFA, HUMBI, and 3DDFA+HUMBI. For each training, we minimize the loss of the reprojection error with weak perspective projection model. To measure the accuracy, we use the reprojection error scaled to the input image resolution (256$\times$ 256 pixels). Table~\ref{table:face_experiment} summarizes the face reconstruction results. The prediction accuracy of 3DDFA+HUMBI is improved from both datasets (2.8 pixels from 3DDFA and 4.9 pixels from HUMBI). which indicates the complementary nature of HUMBI. This indicates that HUMBI is highly complementary to other datasets as it
provides various appearance from 107 viewpoints as shown in Fig.~\ref{fig:view_aug_face}. The qualitative comparisons are shown in Fig.~\ref{fig:face_res}.

\begin{figure}[t]
	\begin{center}
\hspace{-4mm}\includegraphics[width=0.45\textwidth]{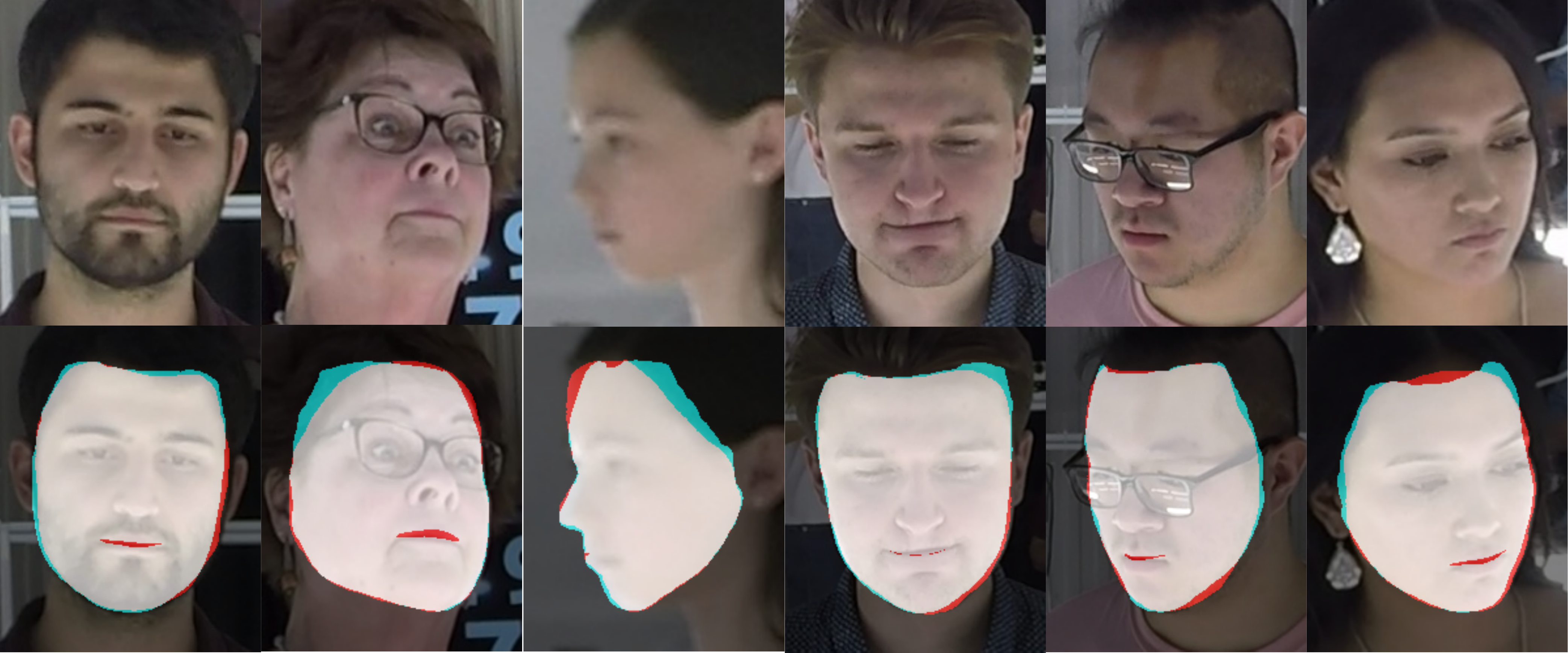}
	\end{center}
	\vspace{-3mm}
    \caption{Silhouette of the reconstructed 3D face overlayed with ground truth. The model is visualized with the blue, the ground truth red, and the overlap white. The ground truth mask is created by manual annotation.}
	\label{fig:acc_humbiface}
\end{figure}

\noindent\textbf{Reconstruction Accuracy} We evaluate how well the reconstructed 3D face model matches the actor by comparing the overlap with the ground truth 2D silhouette. The 3D models of 10 subjects are used for evaluation. We use two metrics: Intersection over Union (IoU) and Chamfer distance between the ground truth mask and the 2D projections of the reconstructed 3D model. In Table~\ref{table:accuracy}, on average, HUMBI Face has 90.9 $\%$ IoU accuracy and 3 pixel distance from the ground truth. The qualitative results are shown in Fig.~\ref{fig:acc_humbiface}.

\begin{table}[t]
\vspace{3mm}
\setlength{\extrarowheight}{2pt}
\centering
\footnotesize
\begin{tabular}{l|c|c}
\hline
 &  IoU & Chamfer distance\\
\hline
HUMBI Face &0.909$\pm$0.021 & 3.009$\pm$0.809   \\
\hline
HUMBI Hand &0.818$\pm$0.036 & 3.340$\pm$1.188   \\
\hline
HUMBI Body &0.820$\pm$0.035 & 12.03$\pm$4.392   \\
\hline

\end{tabular}
\caption{The reconstruction accuracy for HUMBI Face, Hand, and Body measured by Intersection over Union (IoU) and Chamfer distance (unit: pixel) between the ground truth and the projections of the 3D model. HUMBI Body shows the highest Chamfer distance error due to the shape mismatch between 3D body model and dressed human.}
\label{table:accuracy}
\end{table}

\subsection{Hand}
\noindent\textbf{Baseline Datasets} We use three benchmark datasets: (1) Rendered Handpose Dataset (RHD)~\cite{zimmermann2017learning} is a synthesized hand dataset containing 44K images built from 20 freely available 3D models performing 39 actions.  (2) Stereo Hand Pose Tracking Benchmark (SHPTB)~\cite{zhang2017hand} is a real hand dataset captured by a stereo camera rig. (3) FreiHAND~\cite{Freihand2019} is a multiview real hand dataset captured by 8 cameras. (4) ObMan~\cite{hasson19_obman} is a large scale synthetic hand mesh dataset with associated 2D images (141K pairs). We use SHPTB, RHD, and FeiHand datasets for the hand keypoint evaluation and ObMan dataset for the hand mesh evaluation.

\begin{table}[t]
\setlength{\extrarowheight}{2pt}
\centering
\scriptsize
\begin{tabular}{l||c|c|c|c|c|c|c}
\hline
\backslashbox[20mm]{Testing}{Training}&S &R &F &H &S+H &R+H &F+H\\
\hline
SHPTB~\cite{zhang2017hand} (S)  &0.72 &0.40 &0.22 &0.47 &0.40 &0.52 &0.44\\
\hline
RHD~\cite{zimmermann2017learning} (R) &0.16 &0.59 &0.26 &0.49 &0.48 &0.50 &0.44 \\
\hline
FreiHand~\cite{Freihand2019} (F) &0.15 &0.40 &0.72 &0.37 &0.35 &0.43 &0.35 \\
\hline
HUMBI (H) &0.16 &0.36 &0.18 &0.50 &0.43 &0.47 &0.41 \\
\hline\hline
Average &0.30 &0.44 &0.36 &\textbf{0.46} &\textbf{0.42} &\textbf{0.48} &\textbf{0.41} \\
\hline
\end{tabular}
\caption{The cross-data evaluation of 3D hand keypoint prediction. AUC of PCK is used for a metric over an error range of 0-20 mm. Overall, training with HUMBI (H) shows the better performance than other datasets, and the combined dataset (R+H) further improves the AUC scores.}
\label{table:handpose_experiment}
\end{table}
\begin{table}[t]
\setlength{\extrarowheight}{2pt}
\centering
\footnotesize
\begin{tabular}{l|c|c|c}
\hline
\backslashbox[20mm]{Testing}{Training}    &  ObMan & HUMBI & ObMan+HUMBI \\
\hline
ObMan~\cite{hasson19_obman} & 3.84$\pm$2.6 & 6.1$\pm$4.1 &\textbf{3.5$\pm$2.4} \\
\hline
HUMBI & 10.6$\pm$11.3 & 6.5$\pm$8.4 & \textbf{4.8$\pm$5.8}  \\ 
\hline
\end{tabular}
\caption{The mean error of 3D hand mesh prediction for cross-data evaluation (unit: pixel). HUMBI helps with mitigating the domain bias from the synthetic dataset (ObMan), leading to the better reconstruction accuracy.}
\label{table:hand_experiment}
\end{table}

\begin{figure}[t]
	\begin{center}
		\includegraphics[width=0.5\textwidth]{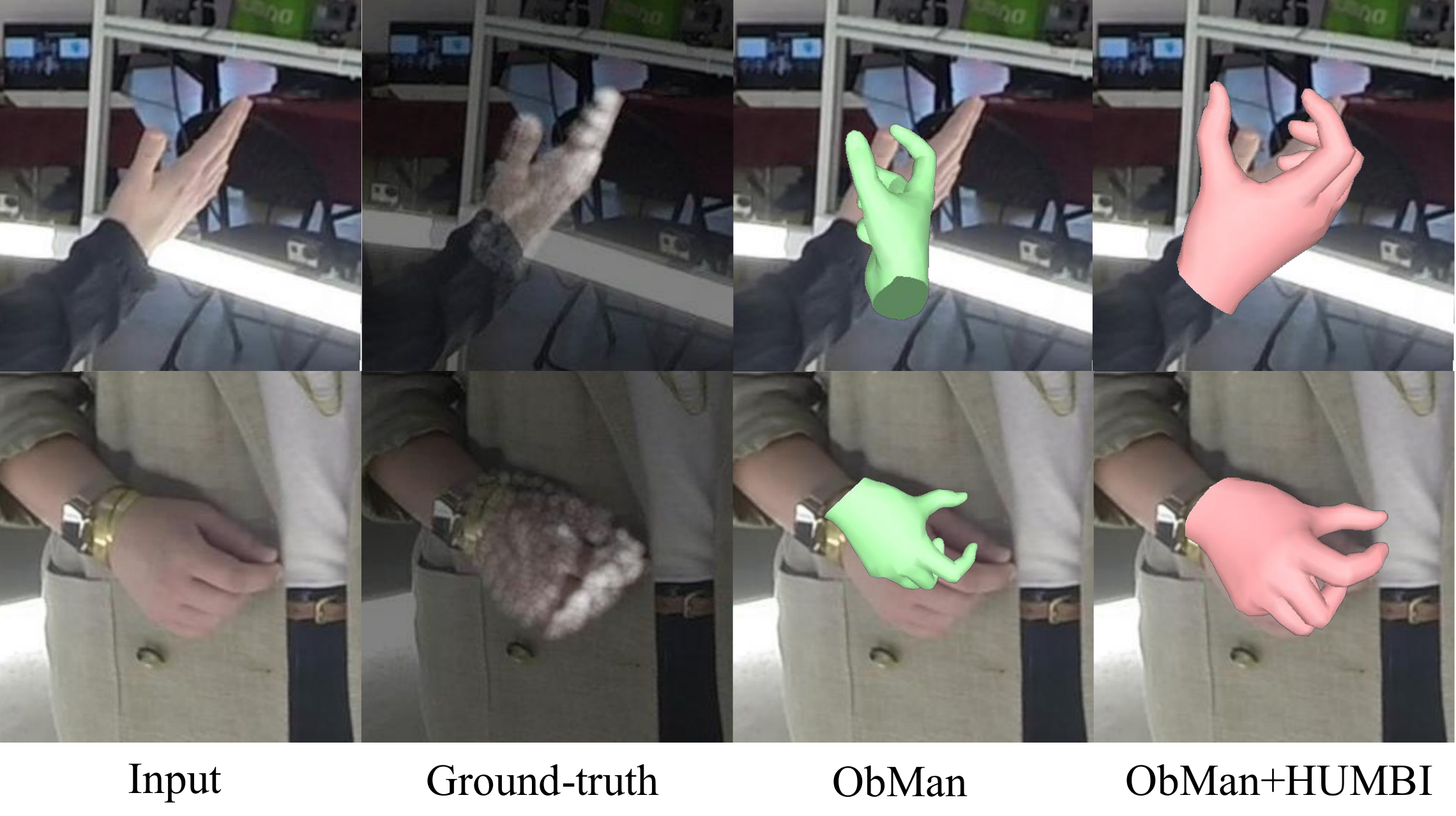}
	\end{center}
	\vspace{-5mm}
    \caption{ Monocular 3D hand mesh prediction results tested on HUMBI Hand. ObMan~\cite{hasson19_obman} is the synthetic dataset.}
    \label{fig:hand_res}
\end{figure}

\noindent\textbf{Monocular 3D Hand Pose Prediction}
We conduct a cross-data evaluation for the task of the 3D hand pose estimation from a single view image using a vanilla hand pose detector~\cite{zimmermann2017learning}.
We train and evaluate the model trained on each dataset and the one combined with HUMBI Hand. The results are summarized in Table~\ref{table:handpose_experiment}. We use the area under PCK curve (AUC) as a metric. HUMBI Hand shows superior performance on predicting 3D hand pose comparing to other three datasets by a margin of 0.02-0.16 AUC. Moreover, HUMBI is complementary to the other datasets, i.e., combining with HUMBI produces more accurate prediction with a margin of 0.04-0.12 AUC.

\noindent\textbf{Monocular 3D Hand Mesh Prediction}
We compare HUMBI Hand with synthetic ObMan~\cite{hasson19_obman} dataset for a mesh reconstruction task on the MANO model~\cite{MANO:SIGGRAPHASIA:2017}. Similar to the face mesh reconstruction Sec.~\ref{exp:face}, we use the image-encoder and mesh-decoder design to form the vanilla network~\cite{Yoon_2019_CVPR}. 
We train and evaluate the network based on the reprojection error with weak perspective projection model. The results are summarized in Table~\ref{table:hand_experiment}. Due to the domain gap between the real and synthetic data, the mesh reconstruction network trained on the synthetic data exhibits limited performance on real HUMBI data while the network trained on HUMBI performs equally. Further, by combining two datasets, the performance is highly improved (even better than intra-data evaluation), e.g., ObMan+HUMBI outperforms ObMan and HUMBI 0.3 and 1.7 pixels, respectively, as shown in Fig.~\ref{fig:hand_res}.

\begin{figure}[t]
	\begin{center}
\hspace{-4mm}\includegraphics[width=0.45\textwidth]{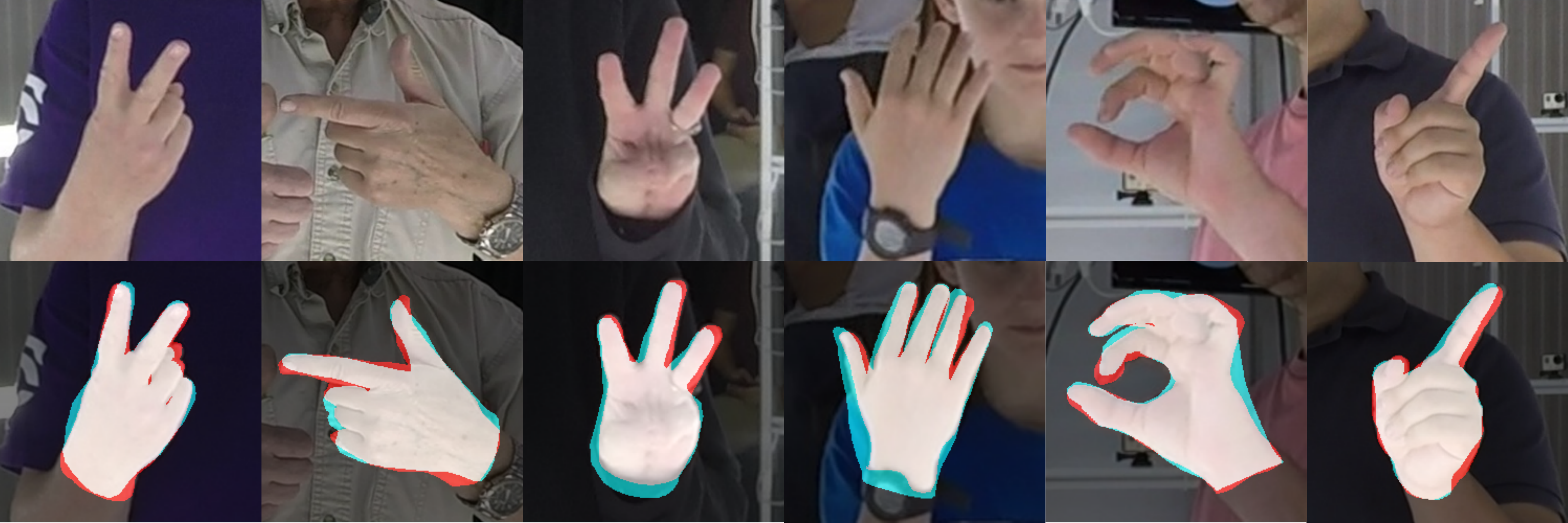}
	\end{center}
	\vspace{-3mm}
    \caption{Silhouette of the reconstructed 3D hand overlayed with ground truth. The model is visualized with the blue, the ground truth red, and the overlap white.}
	\label{fig:humbi_hand_acc}
\end{figure}

\noindent\textbf{Reconstruction Accuracy} We evaluate the accuracy for the reconstructed 3D hand model by comparing the overlap with the ground truth silhouette. The 3D models of 10 subjects are used for evaluation. Similar to HUMBI Face, we use IoU and Chamfer distance to measure the accuracy. In Table~\ref{table:accuracy}, on average, HUMBI Hand has 81 $\%$ IoU accuracy, and 3.3 pixel distance from the ground truth silhouette. The qualitative results are shown in Fig.~\ref{fig:humbi_hand_acc}.



\subsection{Body}
\noindent\textbf{Baseline Datasets} We use four baseline datasets: (1) Human3.6M (H36M) ~\cite{h36m_pami} contains numerous 3D human poses of 11 actors/actresses measured by motion capture system with corresponding images from 4 cameras. (2) MPI-INF-3DHP (MI3D)~\cite{mehta2017monocular} is a 3D human pose estimation dataset which is composed of images with 2D and 3D pose labels captured in both indoor and outdoor scenes. We use its test set containing 2,929 valid images from 6 subjects. 
(3) UP-3D~\cite{lassner2017unite} is a 3D body mesh dataset including 9K images with 3D body meshes. We use Human3.6M and MPI-INF-3DHP for body pose evaluation and UP-3D for body mesh evaluation.

\begin{figure}[t]
	\begin{center}
		\includegraphics[width=0.5\textwidth]{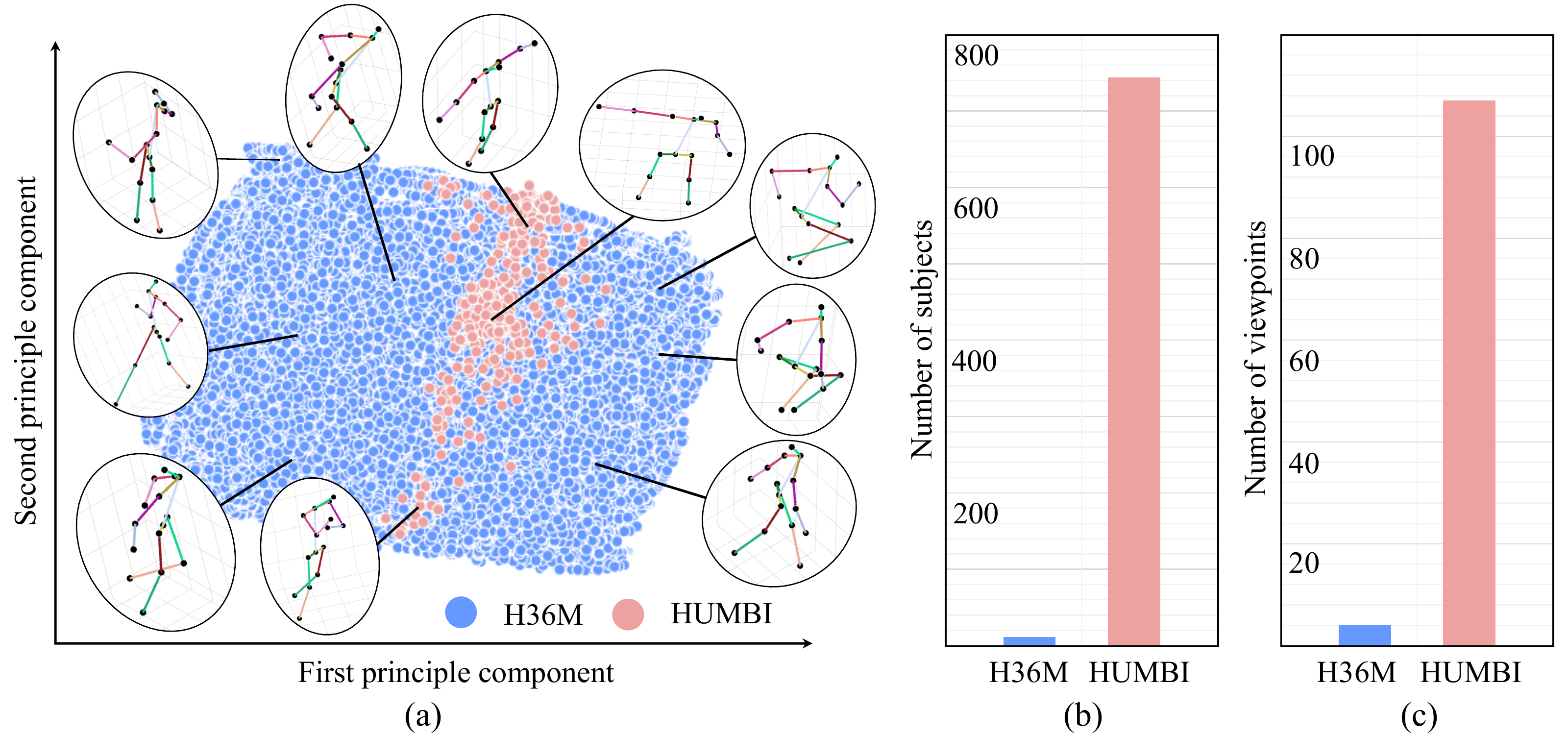}
	\end{center}
	\vspace{-5mm}
    \caption{The comparison of the dataset distribution between Human3.6M (H36M)~\cite{h36m_pami} and HUMBI Body. (a) The distribution of the 3D poses \textit{per subject} in each dataset. We visualize the first and second principal components of the normalized 3D poses where each joint is represented by unit vectors. (b) The number of subjects in each dataset. (c) The number of camera viewpoints in each dataset.}
    \label{fig:data_dist}
\end{figure}

\noindent\textbf{Monocular 3D Body Pose Prediction}
We conduct a cross-data evaluation for the task of estimating 3D human pose from a single view image. We use the 3D body pose detector~\cite{zhou2017towards} as a base network.
We train the model on each dataset alone as well as a mix of HUMBI Body and each of the other two datasets. We evaluate the resulting models on each of those 3 datasets. We use 2D landmark labels from MPII dataset~\cite{andriluka14cvpr} as a weak supervision similar to the training scheme of \cite{zhou2017towards}. The results are summarized in Table~\ref{table:body_experiment}. We use the area under PCK curve (AUC) in an error range of 0-150 mm as the metric. HUMBI shows superior performance on predicting 3D body pose comparing to Human3.6M and MPI-INF-3DHP with a margin of 0.023 and 0.064 AUC. Moreover, HUMBI is complementary to each dataset, i.e., the performance of the model trained by another dataset is increased (by a margin of 0.057 and 0.078 AUC, respectively. We further demonstrate the complementary nature of HUMBI by comparing the pose distribution of HUMBI and Human3.6M (H36M)~\cite{h36m_pami} as shown in Fig.~\ref{fig:data_dist}. H36M provides assorted 3D poses per subject, e.g., HUMBI does not include sitting poses, while HUMBI provides the appearance of diverse subjects seen from a number of viewpoints.

\begin{table}[t]
\setlength{\extrarowheight}{2pt}
\centering
\scriptsize
\begin{tabular}{l||c|c|c|c|c}
\hline
\multirow{2}{*}{\backslashbox[20mm]{Testing}{Training}} 
& \multirow{2}{*}{H36M} & \multirow{2}{*}{MI3D} & \multirow{2}{*}{HUMBI} & H36M &MI3D\\
& & & & +HUMBI & +HUMBI \\
\hline
H36M~\cite{h36m_pami}  &0.562 &0.362 &0.434 &0.551 &0.437\\
\hline
MI3D~\cite{mehta2017monocular} &0.317 &0.377 &0.354 &0.375 &0.425 \\ 
\hline
HUMBI &0.248 &0.267 &0.409 &0.372 &0.377 \\
\hline\hline
Average &0.376 &0.335 &\textbf{0.399} &\textbf{0.433} &\textbf{0.413} \\
\hline
\end{tabular}
\caption{The cross-data evaluation of 3D body keypoint prediction. AUC of PCK is used for a metric over an error range of 0-150 mm. On average, the model trained with HUMBI shows the better performance than the ones with other datasets, and this performance is further improved by mixing them together.}
\label{table:body_experiment}
\end{table}

\begin{table}[t]
\setlength{\extrarowheight}{2pt}
\centering
\footnotesize
\begin{tabular}{l|c|c|c}
\hline
\backslashbox[20mm]{Testing}{Training}    &  UP-3D & HUMBI & UP-3D+HUMBI \\
\hline
UP-3D~\cite{lassner2017unite} & 22.7$\pm$18.6 &49.4$\pm$0.09 &\textbf{18.4$\pm$13.8} \\
\hline
HUMBI & 26.0$\pm$19.7 & 14.5$\pm$6.6 & \textbf{12.5$\pm$8.4}  \\ 
\hline
\end{tabular}
\caption{The mean error of 3D body mesh prediction for cross-data evaluation (unit: pixel) where learning with the combined dataset shows the best performance.}
\label{table:body_experiment1}
\end{table}

\begin{figure}[t]
	\begin{center}
		\includegraphics[width=0.5\textwidth]{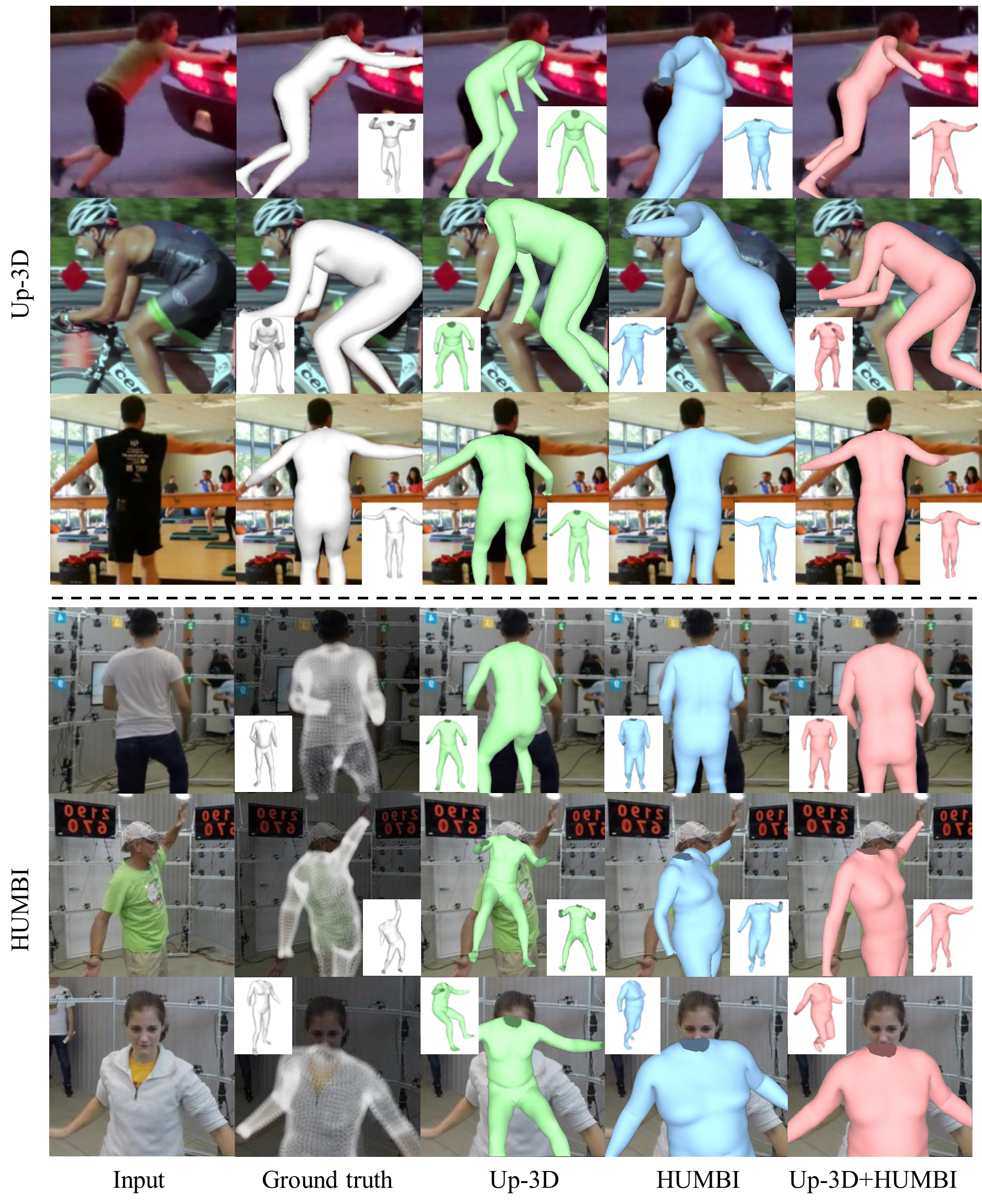}
	\end{center}
	\vspace{-5mm}
    \caption{The qualitative results of the monocular 3D body prediction network trained on different data combination. The column and row represent the type of training and testing data, respectively.
    }
    \label{fig:smpl2}
        \end{figure}

\begin{figure}[t]
	\begin{center}
\hspace{-4mm}\includegraphics[width=0.4\textwidth]{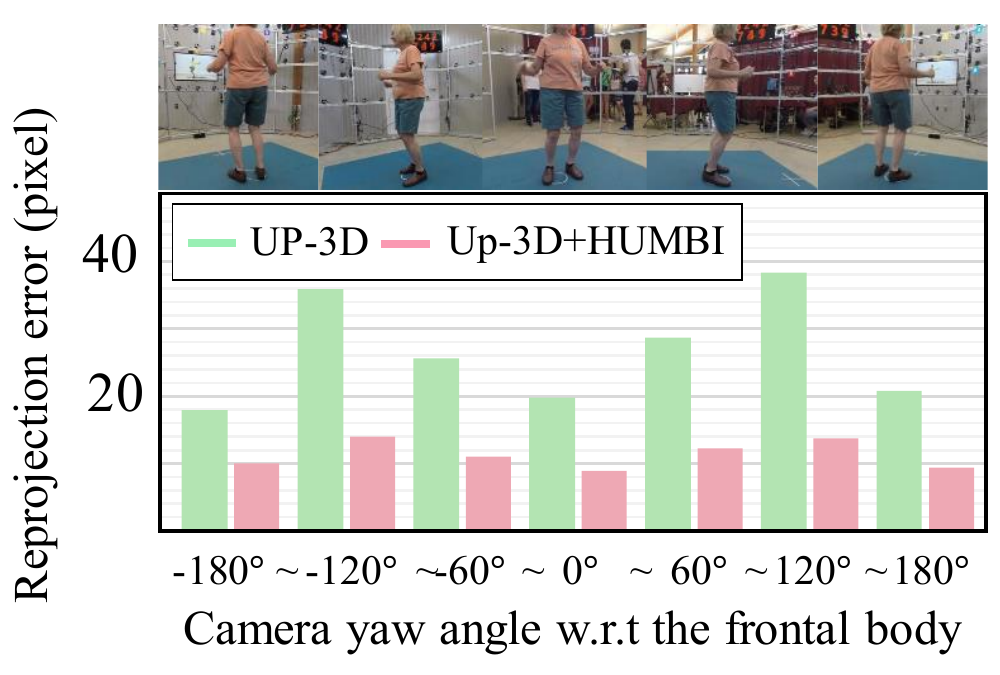}
	\end{center}
	\vspace{-3mm}
    \caption{We measure the viewpoint dependency of body mesh reconstruction models. Combining with HUMBI enforces learning a representation agnostic to viewpoints.} 
    \label{fig:view_aug_body}
\end{figure}


\noindent\textbf{Monocular 3D Body Mesh Prediction}
We compare the body mesh prediction accuracy using the vanilla network model (similar to face mesh reconstruction in Sec~\ref{exp:face}) trained on (1) HUMBI, (2) UP-3D, and (3) HUMBI+UP-3D. The mesh decoder generates SMPL parameters, and the camera pose decoder estimate the camera extrinsic parameters. We train these two decoders by minimizing the reprojection error with the multiview annotations. The cross-data evaluation is summarized in Table~\ref{table:body_experiment1} and the associated qualitative comparisons are shown in Fig.~\ref{fig:smpl2}. We observe that the network trained with HUMBI shows weak performance due to the lack of diversity in poses. However, the performance of the model trained by the combined datasets (i.e., HUMBI+UP-3D) shows an increase of 2 pixels from the model trained by HUMBI alone and 4.3 pixels from Up-3D, indicating that HUMBI is highly complementary to the other datasets. Further, Fig.~\ref{fig:view_aug_body} shows that HUMBI is effective to alleviate the viewpoint bias of the existing dataset.

\begin{figure}[t]
	\begin{center}
\hspace{-4mm}\includegraphics[width=0.45\textwidth]{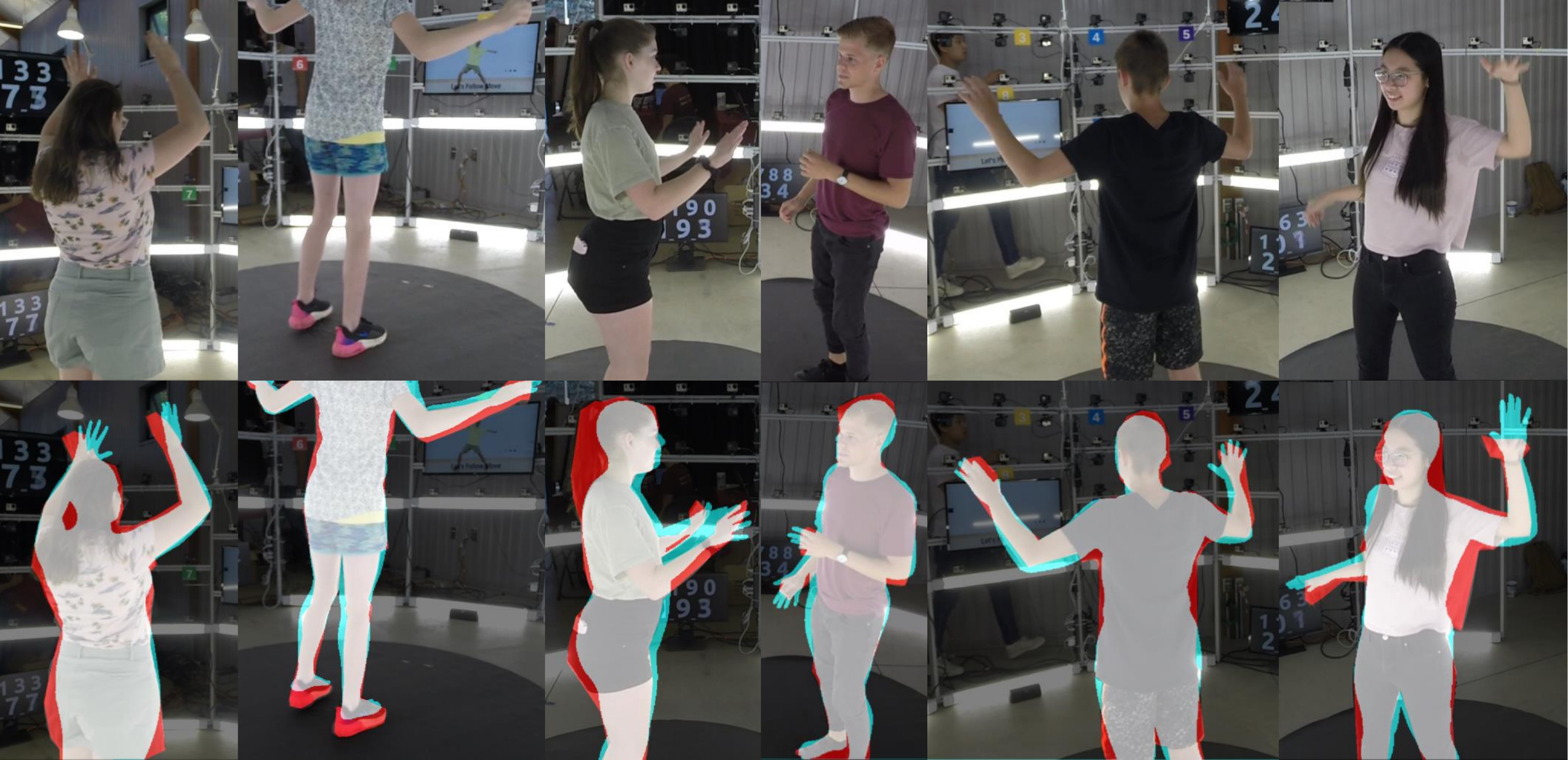}
	\end{center}
	\vspace{-3mm}
    \caption{Silhouette of the reconstructed 3D body overlayed with ground truth. The model is visualized with the blue, the ground truth red, and the overlap white.}
    \label{fig:acc_humbibody}
\end{figure}

\noindent\textbf{Reconstruction Accuracy} We evaluate the accuracy of HUMBI Body by comparing the overlap with the ground truth silhouette. The 3D models of 10 subjects are used for evaluation. Similar to HUMBI Face, we use IoU and Chamfer distance to measure the accuracy. In Table~\ref{table:accuracy}, HUMBI Body has 82 $\%$ IoU accuracy and 12.03 pixel distance from the ground truth. The qualitative results are shown in Fig.~\ref{fig:acc_humbibody}.


{
\renewcommand{\tabcolsep}{4pt}
\begin{table}[t]
\setlength{\extrarowheight}{2pt}
\centering
\scriptsize
\begin{tabular}{l|c|c|c}
\hline
\backslashbox[15mm]{Type}{Style}    &  Short (IoU/Chamf) & Half (IoU/Chamf) & Long (IoU/Chamf) \\
\hline
Top & 0.73 / 10.9 & 0.90 / 5.10 &0.85 / 7.38\\
\hline
Bottom & 0.86 / 6.38 & 0.83 / 9.07 & 0.87 / 6.27  \\ 
\hline
\end{tabular}
\caption{The summary of the garment reconstruction accuracy. We measure the accuracy with the Intersection over Union (IoU) and Chamfer distance (unit: pixel) between the ground truth and the reprojection of the 3D garment.}
\label{table:garment}
\vspace{-2mm}
\end{table}
}

\begin{figure}[h]
	\begin{center}
		\includegraphics[width=3.5in]{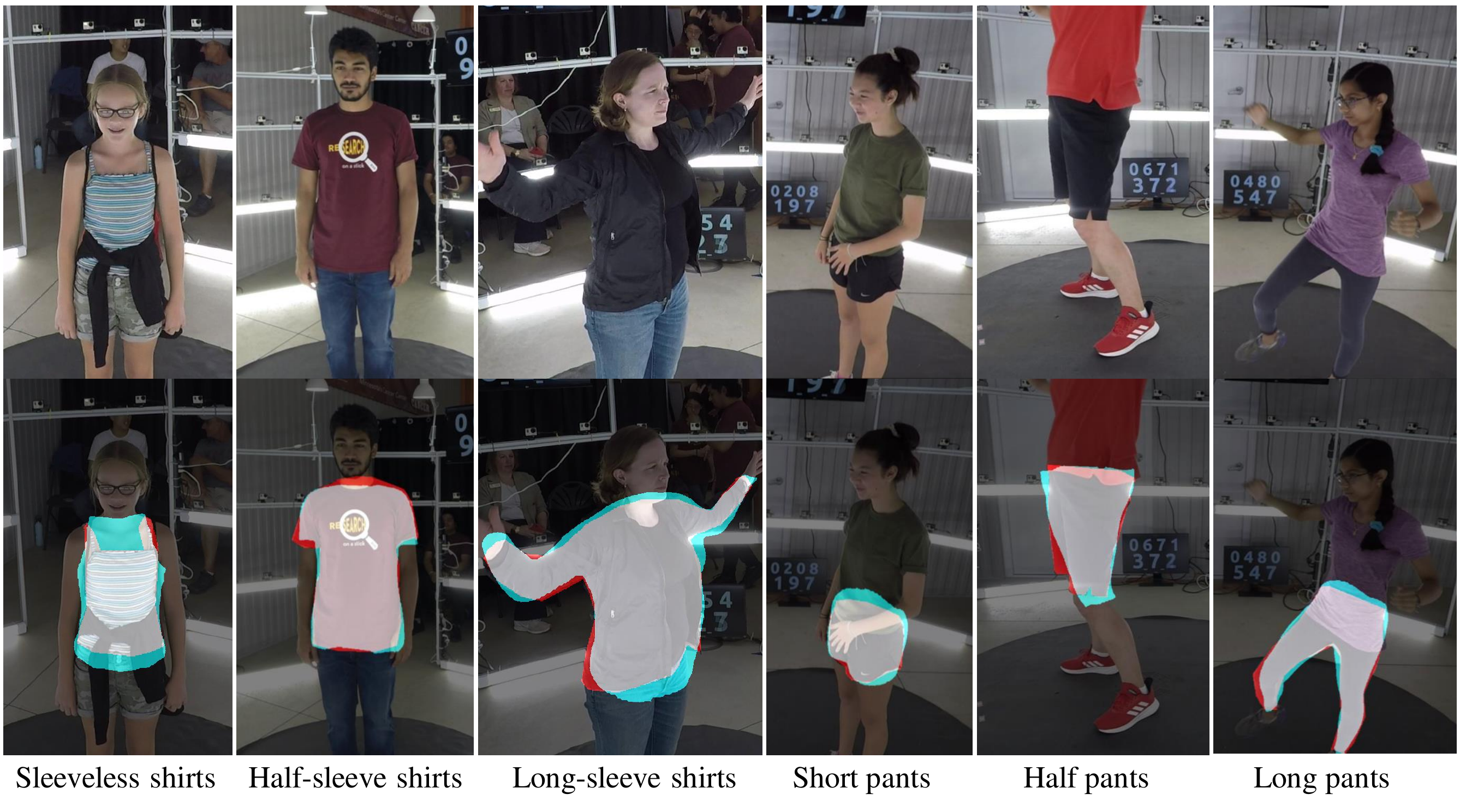}
	\end{center}
	\vspace{-3mm}
    \caption{Silhouette of the reconstructed 3D garments overlayed with ground truth. The model is visualized with the blue, the ground truth with red, and the overlap with white.}
    \label{fig:garment_mask}
\end{figure}

\begin{figure}[h]
	\begin{center}
		\includegraphics[width=3in]{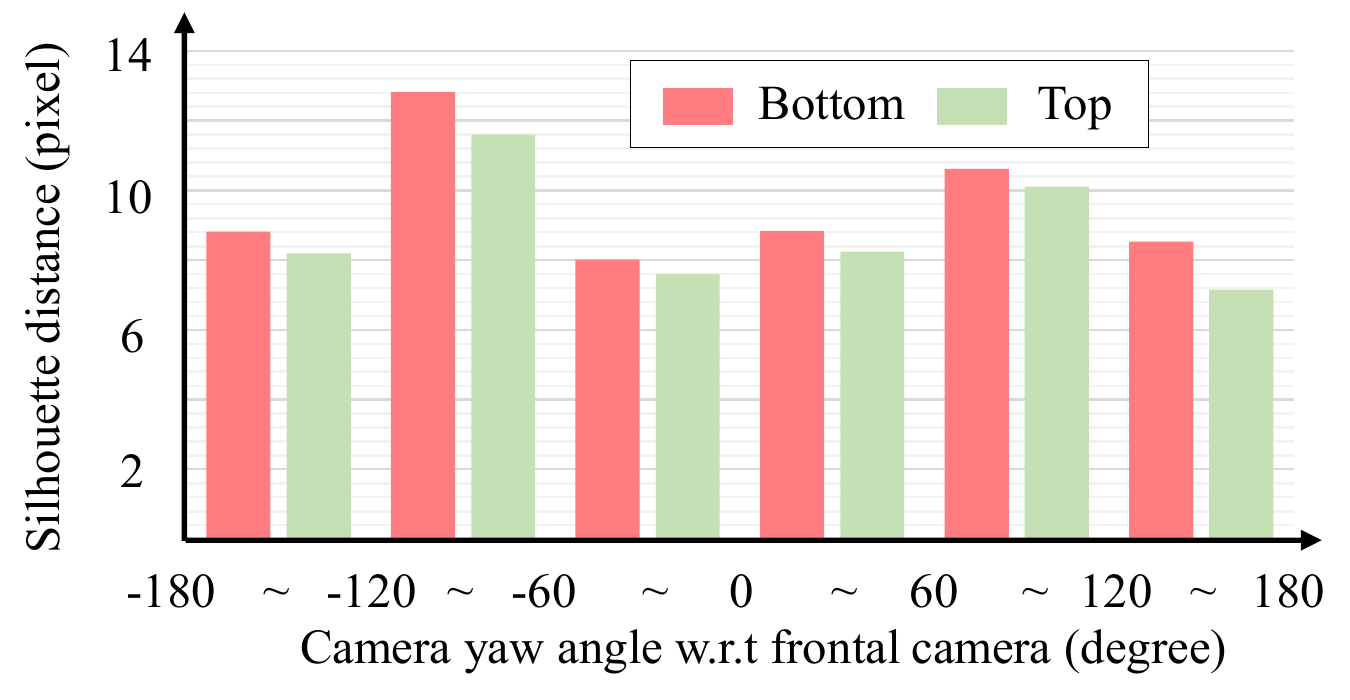}
	\end{center}
	\vspace{-3mm}
    \caption{Garment silhouette error.}
    \label{fig:garment}
	\vspace{-5mm}
\end{figure}

\subsection{Garment}
Unlike other body expressions where common mesh models are used across datasets, garments are dataset-specific. Each dataset has different cloth mesh topology and type, which makes cross-dataset evaluation difficult without non-trivial modifications/augmentations. Instead, we use a topology agnostic metric using IoU of the projected silhouette across views, which can approximate 3D shape evaluation. 
To measure the geometric accuracy, we use two metrics: Intersection over Union (IoU) and Chamfer distance between the ground truth mask and the one from the 2D reprojection of the reconstructed 3D garment mesh. We manually segment the ground truth region using an interactive segmentation tool~\cite{rother2012interactive} considering the occlusion. We subsampled five subjects for each garment model (sleeveless shirts, half-sleeve shirts, long-sleeve shirts, short pants, half pants, and long pants as introduced in Sec.~\ref{humbi:cloth}) and report the mean accuracy in Table~\ref{table:garment}. Overall, the geometric accuracy of our clothing reconstruction shows more than 0.84 overlap ratio with the ground truth mask and less than 7.51 pixel distance from the annotated garment boundary on average. In Fig.~\ref{fig:garment_mask}, we visualize the silhouette of the reconstructed 3D garment overlayed with the ground truth. In addition, we provide the evaluation on the view-dependency of the garment reconstruction. For this, we pick a half-sleeve shirts and half pants models as a representative garment of top and bottom and measure the accuracy based on the Chamfer distance from each camera view that has different angle with respect to the most frontal camera.  On average shown in Fig.~\ref{fig:garment}, the silhouette error seen from the side view (11 pixels) is higher than the frontal (7.5 pixels) and rear views (8 pixels).

\section{Benchmark Challenge}
HUMBI provides multiview images of many subjects with diverse poses, which offers a unique opportunity to evaluate a task of human appearance rendering.  We formulate a new benchmark challenge of rendering that can facilitate photo-realistic human rendering research.







\noindent\textbf{Task Definition} Given an image of a person and a target pose, render the appearance of the person that agrees with the target pose.


\noindent\textbf{Benchmark Dataset} We randomly select the pair of the reference and target poses across the time and views. In our experiments, 101K pairs of the views are selected from 100 subjects for training, and 15,923 from 40 subjects for testing. For each view, we use the person image with 256$\times$256 pixel resolution after cropping and resizing based on the projected \textit{xy}-coordinate of 3D keypoints.

\noindent\textbf{Baselines} We evaluate the following six state-of-the-art approaches. \textbf{PG}~\cite{ma2017pose} synthesizes a person image with two generators in coarse-to-fine manner: \textbf{PG-1} and \textbf{PG-2} indicates coarse and fine syntheses. \textbf{C2GAN}~\cite{tang2019cycle} uses the cycle consistency on body keypoints and multiview images. \textbf{PPA}~\cite{zhu2019progressive} integrates the pose attention module into a generative network to progressively refine the image, and \textbf{SGAN}~\cite{tang2020multi} selectively combines multi-channel attention map that enhances the quality of the generated images. For \textbf{PPA}, we use the image with 256$\times$176 pixel resolution. \textbf{GFLA}~\cite{ren2020deep} estimates a global flow field to transform the local attention features. \textbf{NHRR}~\cite{Sarkar2020} warps the pixels from the input image to the target image based on the dense correspondences from  a parametric body model~\cite{loper2015smpl}. We train these networks using the training parameters suggested by the authors.


\noindent\textbf{Metric} We measure the quality of the generated images using Learned Perceptual Image Patch Similarity (LPIPS)~\cite{zhang2018unreasonable} and Frechet Inception Distance (FID)~\cite{heusel2017gans}. LPIPS measures the distance between the generated images and ground truth in a feature space, e.g., VGG features. FID measures the realism of the generated images by computing Wassertein-2 distance between the distributions of the generated images and ground truth. To eliminate the influence of background, we mask out the background region by incorporating segmentation~\cite{yolact-iccv2019} to form Mask-LPIPS and Mask-FID.

\begin{figure*}[h]
	\begin{center}
		\includegraphics[width=7in]{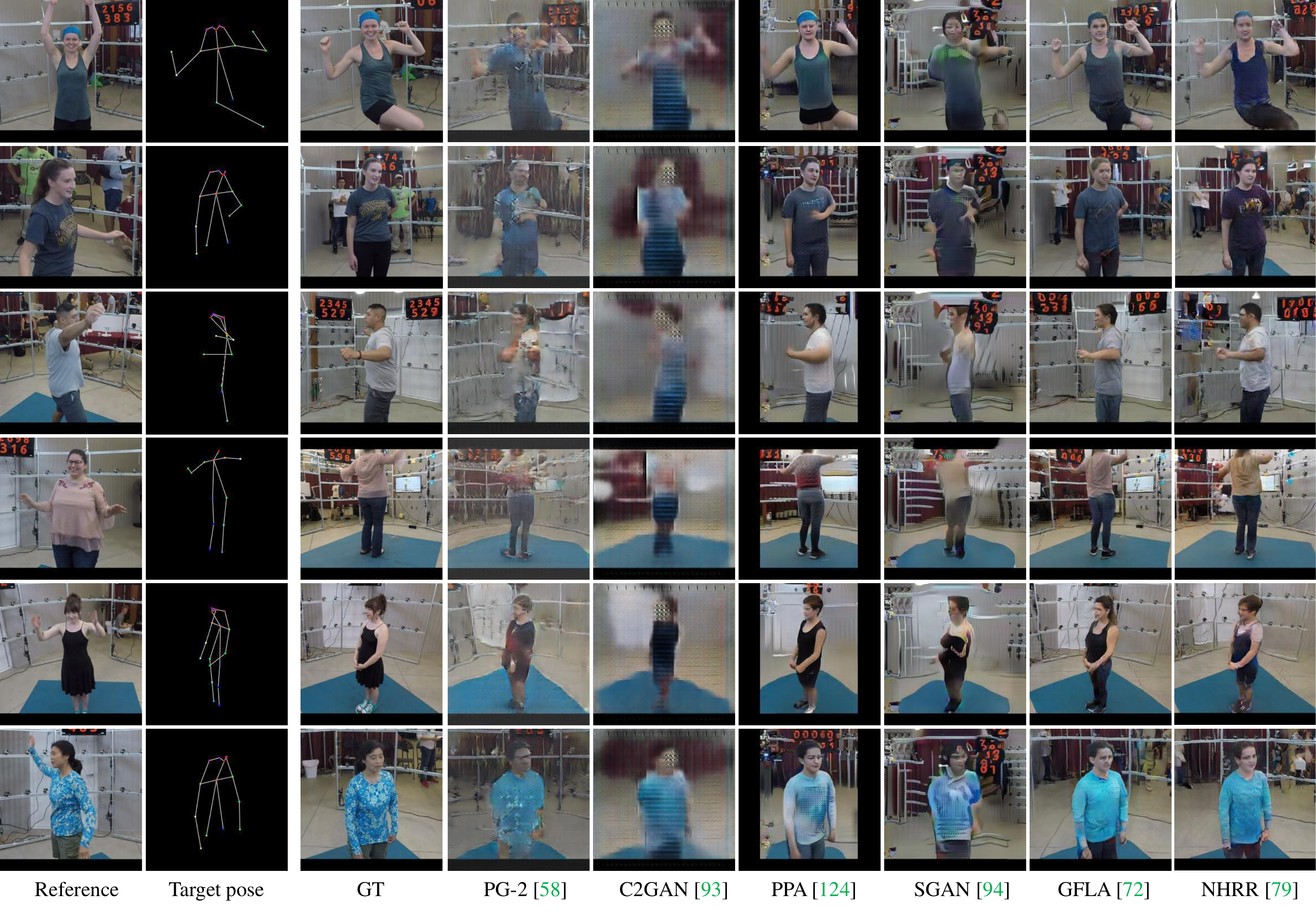}
	\end{center}
	\vspace{-5mm}
    \caption{The qualitative comparison of pose-guided person image generation from each method. For NHRR~\cite{Sarkar2020}, the densepose detection~\cite{guler2018densepose} is used as a conditioning target pose.}
    \label{fig:pose-guided}
	\vspace{-5mm}
\end{figure*}

\noindent\textbf{Analysis} Table~\ref{table:pose-guided} summarizes the quantitative evaluations on HUMBI benchmark dataset. GFLA outperforms other methods. It can effectively model the following view-dependent properties by learning our multiview dataset: 1) The network generates the realistic background scene which is dependent on the camera viewpoint as shown in Fig.~\ref{fig:pose-guided}. 2) The network models the view-dependent lighting, that can be verified in Fig.~\ref{fig:pose-guided}-(third row), e.g., the color of T-shirt is bluish from the source view, while it is grayish from the target view. This indicates that the network can implicitly model the camera viewpoint from the target body pose and decode such view-dependant properties on the generated images. The comparison of FID with Mask-FID shows that the performance of PPA is significantly improved with the human mask, highlighting that it focuses only on the person region NHRR shows the best performance on the Mask-LPIPS metric, i.e., the rendered human images are perceptually close to real.

It is also worth to note several limitations observed from these state-of-the-art approaches. The person specific visual features, e.g., color and shape of the face and hair, in the generated images are not photo-realistically rendered. Transferring  a variety of clothing style is more challenging. For example, the dress in the reference image is transferred to long pants from the generated image as shown in Fig.~\ref{fig:pose-guided}-(fifth row), and a lacy shirt is converted to a tight T-shirt in Fig.~\ref{fig:pose-guided}-(fourth row). They also fail to transfer the clothing textures in a semantically meaningful way, e.g., the flower patterns on the clothing from the reference image in Fig.~\ref{fig:pose-guided}-(sixth row) is not preserved in the generated ones. 
Likewise, human rendering from a single image of diverse subjects is still far behind the metric-level accuracy, i.e., the rendering does not match to the ground truth image at each pixel location, and there exists substantial rooms to improve. This will encourage future research to push the boundary of photorealistic human rendering with tera-scale multiview imaging dataset.


{
\renewcommand{\tabcolsep}{5pt} 
\begin{table}[t]
\vspace{3mm}
\setlength{\extrarowheight}{2pt}
\centering
\scriptsize
\begin{tabular}{l||c|c|c|c}
\hline
\multirow{2}{*}{\backslashbox[20mm]{Baseline}{Metric}} & \multirow{2}{*}{LPIPS} & \multirow{2}{*}{Mask-LPIPS} & \multirow{2}{*}{FID} & \multirow{2}{*}{Mask-FID} \\
& & & & \\
\hline
PG-1~\cite{ma2017pose}  & 0.680$\pm$0.076 & 0.154$\pm$0.048  & 290.45 & 139.44  \\
\hline
PG-2~\cite{ma2017pose}  & 0.465$\pm$0.064 & 0.141$\pm$0.043 & 105.56 & 80.64  \\
\hline
C2GAN~\cite{tang2019cycle} & 0.641$\pm$0.066 & 0.166$\pm$0.048 & 241.44 & 148.55  \\
\hline
PPA~\cite{zhu2019progressive} & 0.537$\pm$0.043 & 0.137$\pm$0.049 & 114.08 & 19.95  \\
\hline
SGAN~\cite{tang2020multi} & 0.525$\pm$0.044 & 0.164$\pm$0.051 & 111.58 & 34.91  \\
\hline
GFLA~\cite{ren2020deep} & \textbf{0.328$\pm$0.090}  & 0.122$\pm$0.042 & \textbf{14.25} & \textbf{13.49}  \\
\hline
NHRR~\cite{Sarkar2020} & 0.356$\pm$0.093  & \textbf{0.115$\pm$0.038} & 19.21 &17.49  \\
\hline
\end{tabular}
\caption{The quantitative evaluation of pose-guided person image generation. The lower score shows the better results.}
\label{table:pose-guided}
\end{table}
}

\section{Discussion}
We present HUMBI dataset that is designed to facilitate high resolution pose- and view-specific appearance of human body expressions. Five elementary body expressions (gaze, face, hand, body, and garment) are captured by a dense camera array composed of 107 synchronized cameras. The dataset includes diverse activities of 772 distinctive subjects across gender, ethnicity, age, and physical condition. We use a 3D mesh model to represent the expressions where the view-dependent appearance is coordinated by its canonical atlas. Our evaluation shows that HUMBI outperforms existing datasets as modeling nearly exhaustive views and can be complementary to such datasets. 

HUMBI is the first-of-its-kind dataset that attempts to span the general appearance of assorted people by pushing towards two extremes: views and subjects. This will provide a new opportunity to build a versatile model that generates photorealistic rendering for authentic telepresence. However, the impact of HUMBI will not be limited to appearance modeling, i.e., it can offer a novel multiview benchmark dataset for a stronger and generalizable reconstruction and generation model specific to humans.


\section*{Acknowledgement}
This work was partially supported by National Science Foundation (No.1846031 and 1919965) and Institute for Information $\&$ communications Technology Promotion(IITP) grant funded by the Korea government(MSIP) (No. 2019-0-01906, Artificial Intelligence Graduate School Program (POSTECH)).

\ifCLASSOPTIONcaptionsoff
  \newpage
\fi

\bibliographystyle{ieee}
\bibliography{main}



%




%
\vspace{50mm}
\begin{IEEEbiography}[{\includegraphics[width=1in,height=1.25in,clip]{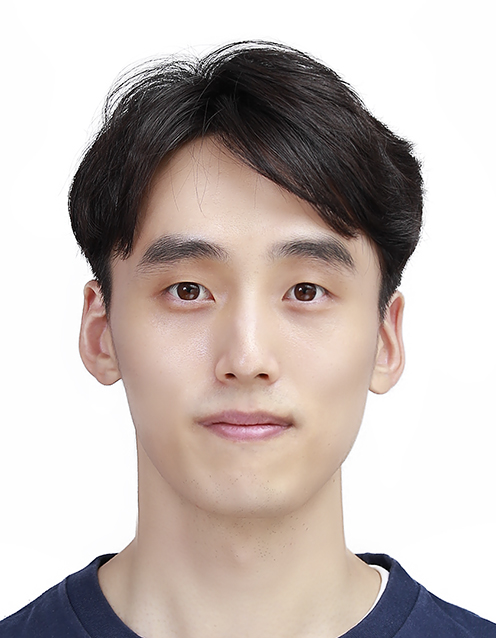}}]{Jae Shin Yoon} is a Ph.D. student in Computer Science and Engineering at University of Minnesota, Twin Cities. He received the B.S. degree in electric engineering from Hanyang University, Seoul, South Korea, in 2015 and the M.S. degree in robotics program from Korea Advanced Institute of Science and Technology (KAIST), Daejeon, South Korea, in 2017. His research focuses on reconstructing and understanding of human dynamics using computer vision and machine learning.
\end{IEEEbiography}
\begin{IEEEbiography}[{\includegraphics[width=1in,height=1.25in,clip]{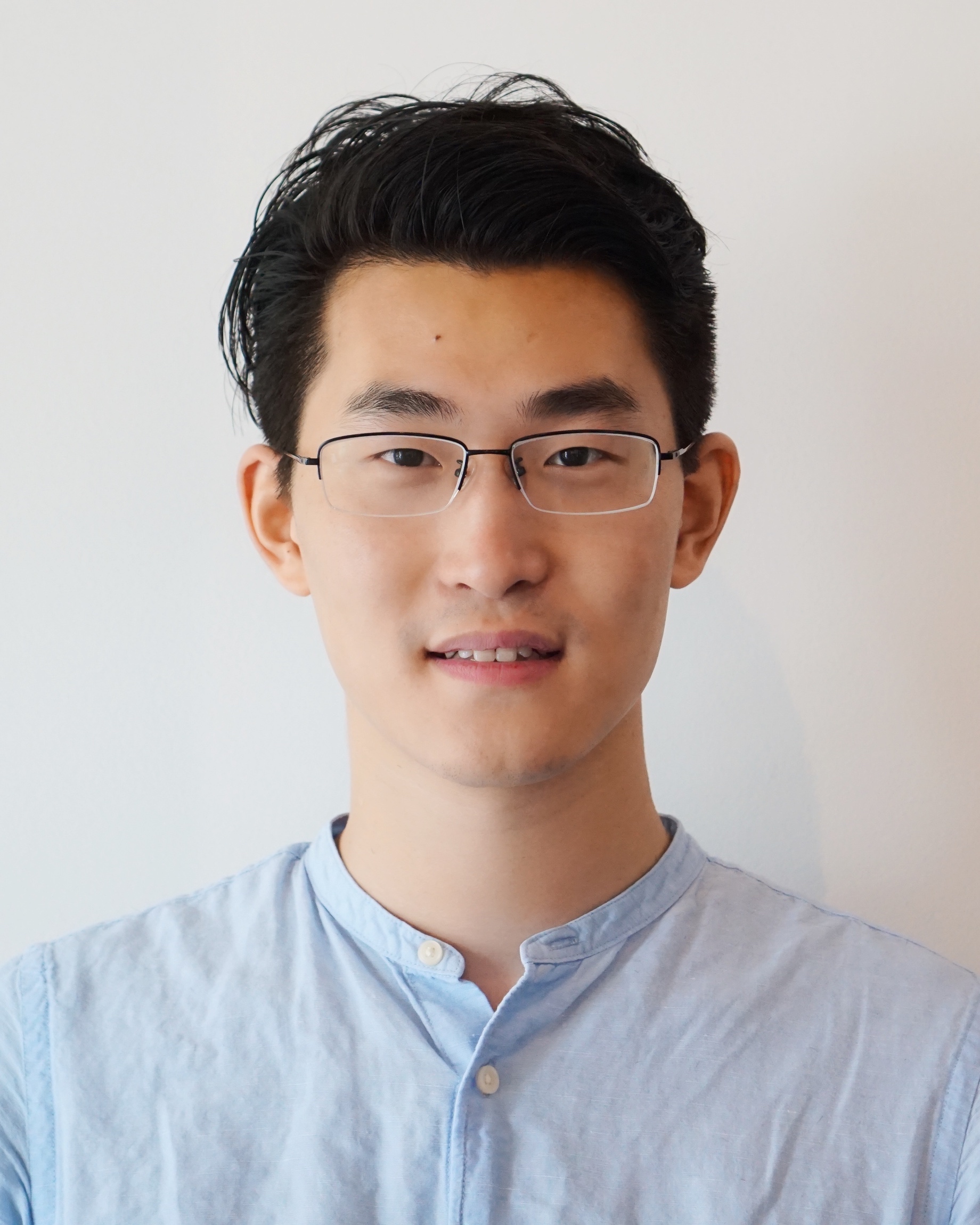}}]{Zhixuan Yu}
Zhixuan Yu is a Ph.D. student in Computer Science and Engineering at University of Minnesota, Twin Cities. He received a B.Eng. degree in biomedical engineering from Huazhong University of Science and Technology in 2015 and a M.S. degree in biomedical engineering from Carnegie Mellon University in 2016. His major research interests are in the area of 3D computer vision with emphasis on human pose estimation.
\end{IEEEbiography}
\vspace{-5mm}
\begin{IEEEbiography}[{\includegraphics[width=1in,height=1.25in,clip]{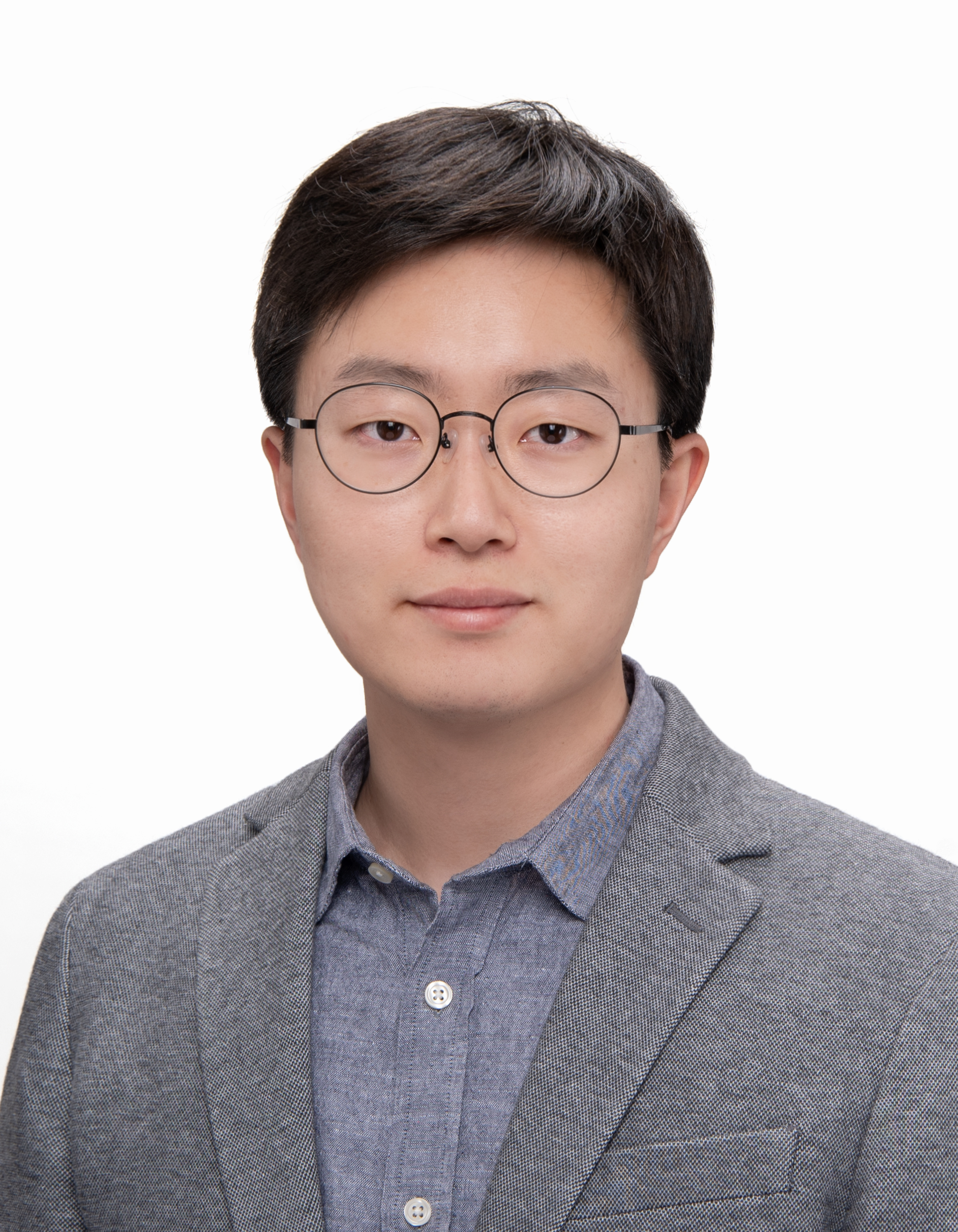}}]{Jaesik Park}
Jaesik Park is an Assistant Professor at POSTECH. He received his Bachelor's degree from Hanyang University in 2009, and he received his Master's degree and Ph.D. degree from KAIST in 2011 and 2015, respectively. He is a recipient of the Microsoft Research Asia Fellowship and two Samsung Humantech Awards. After he joined Intelligent Systems Lab at Intel in 2015 as a Research Scientist, he authored/co-authored academic papers about depth camera-based geometry recovery, an automatic photogrammetry evaluation system, 3D semantic segmentation, and surface registration algorithms. He serves as an Area Chair for ICCV 2019, CVPR 2020, CVPR 2021. 
\end{IEEEbiography}
\vspace{-5mm}
\begin{IEEEbiography}[{\includegraphics[width=1in,height=1.25in,clip]{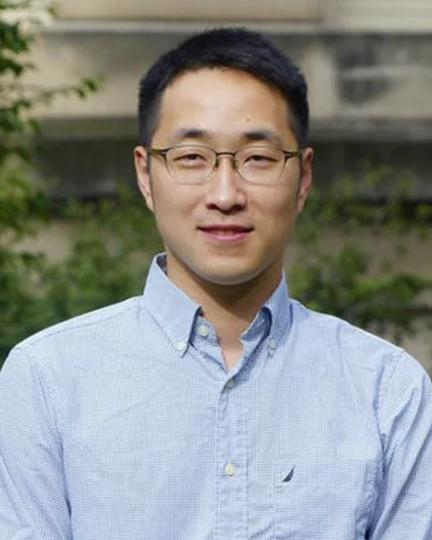}}]{Hyun Soo Park}
Hyun Soo Park is an Assistant Professor at Computer Science and Engineering, the University of Minnesota. He is interested in enabling behavioral imaging using computer vision. He has received NSF's CRII and CAREER awards. He was a Postdoctoral Fellow in GRASP Lab at University of Pennsylvania. He earned his Ph.D. from Carnegie Mellon University.  
\end{IEEEbiography}






\end{document}